\documentclass[conference]{IEEEtran}
\usepackage{times}

\usepackage[numbers]{natbib}
\usepackage{multicol}
\usepackage[bookmarks=true，colorlinks=true]{hyperref}

\usepackage{booktabs}
\usepackage{multirow}

\usepackage{amsmath}
\usepackage{amssymb}

\usepackage[table]{xcolor}  
\usepackage{colortbl}
\definecolor{lightblue}{RGB}{235,244,253}

\usepackage{pifont}
\newcommand{\cmark}{\ding{51}} 
\newcommand{\xmark}{\ding{55}} 

\usepackage{graphicx}

\usepackage{pgfplots}
\pgfplotsset{compat=1.18}

\usepackage[utf8]{inputenc}
\usepackage{tikz}
\usepackage{pgf-pie} 

\usepackage{cuted}
\usepackage{caption}

\usepackage{float}
\usepackage[dvipsnames, svgnames]{xcolor}

\newcommand{\best}[1]{%
  \cellcolor{black!18}{\textbf{#1}}%
}

\newcommand{\second}[1]{%
  \cellcolor{black!8}{\textbf{#1}}%
}

\usepackage{pythonhighlight}

\begin{document}

\title{LDA-1B: Scaling \underline{L}atent \underline{D}ynamics \underline{A}ction Model via Universal Embodied Data Ingestion
\vspace{-5mm}}


\author{
    Jiangran Lyu$^{*1,2}$, Kai Liu$^{*2,3,4}$, Xuheng Zhang$^{*1,2}$, Haoran Liao$^{2,6}$, Yusen Feng$^{1,2}$, Wenxuan Zhu$^{1}$, Tingrui Shen$^{1}$, \\ Jiayi Chen$^{1,2}$, Jiazhao Zhang$^{1,2}$, 
    Yifei Dong$^{1}$, Wenbo Cui$^{2,3,4}$, Senmao Qi$^{2}$, Shuo Wang$^{2}$, Yixin Zheng$^{2,3,4}$, Mi Yan$^{1,2}$,\\ Xuesong Shi$^{2}$, Haoran Li$^{3}$, Dongbin Zhao$^{3}$, 
    Ming-Yu Liu$^{7}$, Zhizheng Zhang$^{2,\dagger}$, Li Yi$^{5,\dagger}$, Yizhou Wang$^{1,\dagger}$, He Wang$^{1,2,\dagger}$
    \vspace{0.1cm}
 \\
    \small $^1$Peking University \quad $^2$Galbot \quad $^3$CASIA 
    \quad $^4$BAAI \quad $^5$Tsinghua University \quad $^6$Sun Yat-sen University \quad $^7$NVIDIA \vspace{0.1cm}
    \\ Code \& Data: 
    {\color{RoyalBlue}\textbf{https://pku-epic.github.io/LDA}}
     $^*$ Equal contribution \quad $^\dagger$ Corresponding authors 
}

\maketitle

\begin{strip}\centering
\vspace{-1.5cm}
\includegraphics[width=1\linewidth]{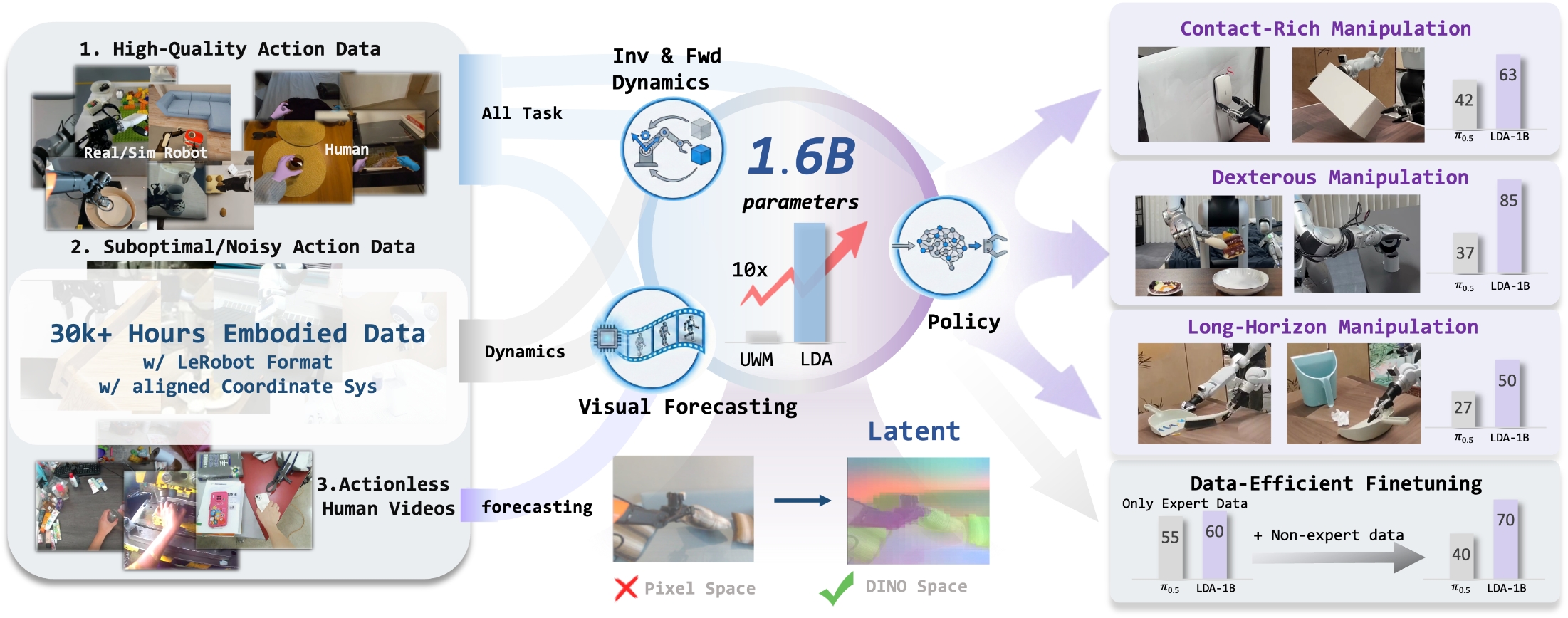}
\vspace{-0.5cm}
\captionof{figure}{We introduce LDA-1B, a 1.6 B-parameter robot foundation model scaled on over 30k hours of heterogeneous embodied data.
LDA-1B unifies policy, dynamics, and visual forecasting in a structured DINO~\cite{simeoni2025dinov3} latent space, allowing different data sources to play complementary roles.
Beyond high-quality data alone, noisy data and actionless videos also provide valuable visual and physical priors for dynamics learning.
This universal data ingestion paradigm enables stable scaling with data and model size, significantly outperforming strong baselines such as $\pi_{0.5}$~\cite{intelligence2025pi_} across diverse manipulation tasks.}

\label{first_figure} 
\vspace{-.2cm}
\end{strip}

\begin{abstract}

Recent robot foundation models largely rely on large-scale behavior cloning, which imitates expert actions but discards transferable dynamics knowledge embedded in heterogeneous embodied data.
While the Unified World Model (UWM) formulation has the potential to leverage such diverse data, existing instantiations struggle to scale to foundation-level due to coarse data usage and fragmented datasets.
We introduce \textbf{LDA-1B}, a robot foundation model that scales through universal embodied data ingestion by jointly learning dynamics, policy, and visual forecasting, assigning distinct roles to data of varying quality.
To support this regime at scale, we assemble and standardize \textbf{EI-30k}, an embodied interaction dataset comprising over \textbf{30k hours} of human and robot trajectories in a unified format.
Scalable dynamics learning over such heterogeneous data is enabled by prediction in a structured DINO latent space, which avoids redundant pixel-space appearance modeling.
Complementing this representation, LDA-1B employs a multimodal diffusion transformer to handle asynchronous vision and action streams, enabling stable training at the \textbf{1B-parameter} scale.
Experiments in simulation and the real world show LDA-1B outperforms prior methods (e.g., $\pi_{0.5}$) by up to \textbf{21\%}, \textbf{48\%}, and \textbf{23\%} on contact-rich, dexterous, and long-horizon tasks, respectively.
Notably, LDA-1B enables data-efficient fine-tuning, gaining 10\% by leveraging 30\% low-quality trajectories typically harmful and discarded.

\end{abstract}

\section{Introduction}

Inspired by the success of Large Language Models (LLMs) and Vision-Language Models (VLMs), the robotics community has increasingly pursued general-purpose robot foundation models through large-scale pretraining~\cite{brohan2022rt, octo_2023}. Most existing approaches center on scaling behavior cloning (BC), which imitates expert actions but fundamentally restricts learning to high-quality demonstrations. 
Consequently, a large portion of heterogeneous embodied data~\cite{padalkar2023open} is discarded or only weakly utilized, despite containing rich physical interaction dynamics~\cite{khazatsky2024droid}.

Unified World Model (UWM) formulation~\cite{li2025unified, zhu2025unified} provides an alternative by jointly optimizing dynamics, policy, and video generation within a single model, which can leverage not only expert data. Despite the potential value, existing UWM instantiations remain far from scaling to foundation-level. 
A major limitation lies in coarse data usage: heterogeneous embodied data are often treated uniformly, without differentiating their roles by quality or supervision, which underutilizes transferable dynamics knowledge. In addition, the community lacks ready-to-use large-scale datasets that unify varying-quality data with consistent formats and aligned action representations.
Furthermore, UWM represents future state in pixel space, entangling dynamics learning with redundant appearance modeling. Subtle variations in illumination, texture, background clutter, or camera viewpoint can dominate the training objective, making large-scale training inefficient and hindering the learning of interaction-relevant dynamics.

To overcome these limitations, we introduce \textbf{LDA-1B}, a robot foundation model that scales via \emph{universal embodied data ingestion}. In this framework, heterogeneous data play distinct yet complementary roles: actionless human videos supervise visual forecasting~\cite{nair2022r3m, ma2022vip, karamcheti2023language}, lower-quality trajectories primarily inform dynamics learning, and high-quality trajectories support both policy and dynamics. To realize this approach at scale, we assemble \textbf{EI-30k}, a large-scale embodied interaction dataset with over 30k hours of human and robot trajectories across real and simulated environments, standardized in format and aligned in action representation. Scalable learning on such diverse data is facilitated by a structured \textbf{DINO latent space}~\cite{simeoni2025dinov3, zhou2024dino, huang2025ladi}, which reduces redundant appearance modeling~\cite{nair2022r3m, karamcheti2023language}, and a multimodal diffusion transformer that aligns asynchronous visual and action prediction. By combining this ingestion strategy, dataset, latent representation, and model architecture, LDA-1B achieves stable training at the 1B-parameter scale while maximizing data utilization.

We evaluate LDA-1B on challenging RoboCasa-GR1 benchmark and a diverse set of real-world tasks involving both grippers and high-DoF dexterous hands~\cite{zhao2023learning}. LDA-1B consistently outperforms $\pi_{0.5}$, achieving \textbf{21\%} gains on contact-rich manipulation, benefiting from improved dynamics understanding and \textbf{48\%} gains on dexterous manipulation, benefiting from effective utilization of human data. Moreover, under a mixed-quality fine-tuning setting, LDA-1B improves data efficiency by \textbf{10\%} through leveraging low-quality trajectories that are detrimental to baseline methods. These results highlight universal embodied data ingestion and unified latent dynamics learning as a scalable alternative to behavior-cloning-centric robot pretraining.
In summary, our contributions are threefold:
\begin{itemize}
    \item We propose LDA-1B, a scalable robot foundation model that learns generalizable interaction dynamics through unified latent dynamics pretraining.
    \item We construct EI-30k, a large-scale embodied interaction dataset covering diverse embodiments, environments, data qualities, with aligned end effector coordinate system.
    \item We demonstrate that LDA-1B achieves superior generalization and robustness across a wide range of settings, including simulation and real-world environments, contact-rich manipulation, dexterous manipulation, and long-horizon manipulation.
\end{itemize}

\begin{figure*}[t]
    \centering
    \includegraphics[width=0.90\linewidth]{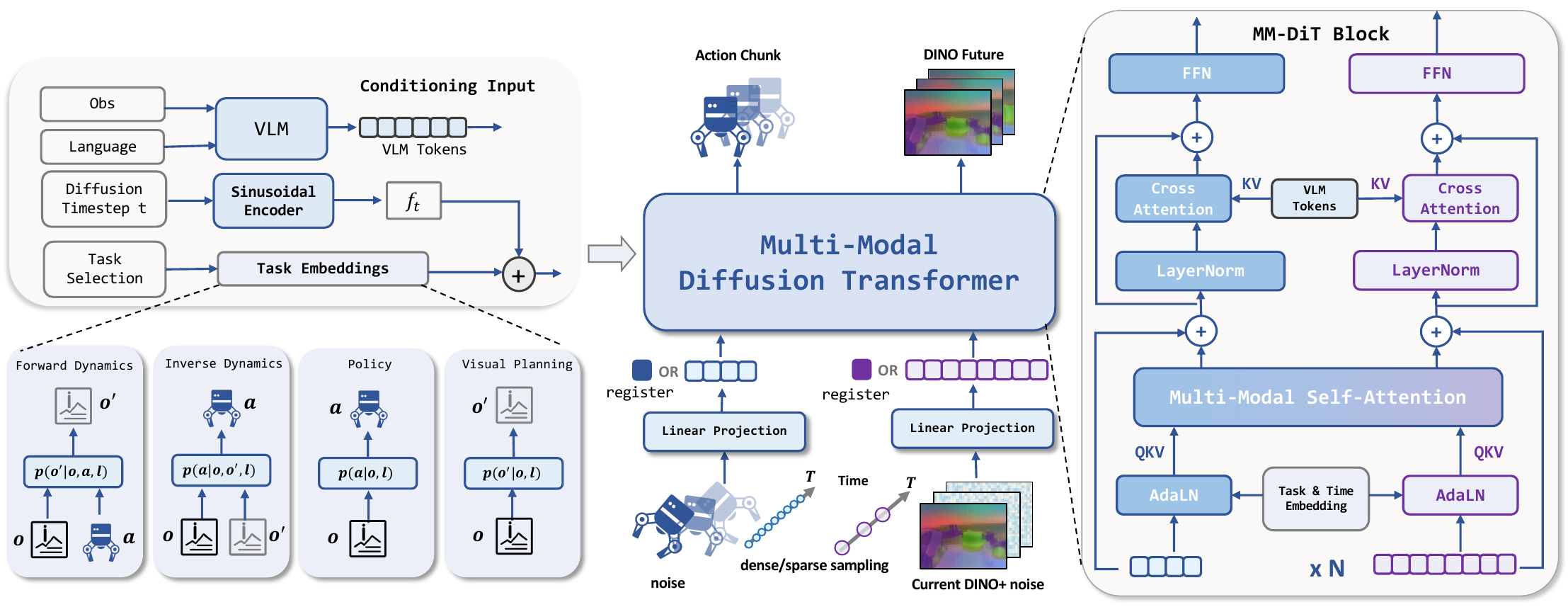}
    \caption{\textbf{Architecture of LDA.}
LDA jointly denoises action chunks and future visual latents under multiple co-training objectives, including policy learning, forward dynamics, inverse dynamics, and visual forecasting.
Conditioned on VLM tokens, diffusion timesteps, and task embeddings, the model adopts a multimodal diffusion transformer architecture, where action and visual experts are decoupled and interact through a shared self-attention layer.}

    \vspace{-2mm}
	\label{fig:pipeline}
\end{figure*}

\section{Related Work}
\setlength{\tabcolsep}{2.1pt}
\begin{table}[t]
    \centering
    \begin{tabular}{lccccc}
        \toprule
        \rowcolor{gray!10}
        Model 
        & Data Src. 
        & \#Data 
        & Action Quality 
        & Train. 
        & Param. \\ 
        \midrule
        $\pi_{0.5}$ \cite{intelligence2025pi_} 
        & Tele. 
        & 10k+ 
        & High 
        & BC 
        & 3B \\

        RDT \cite{liu2024rdt} 
        & Tele. 
        & \textless 10k 
        & High  
        & BC 
        & 1B \\

        GraspVLA \cite{deng2025graspvla} 
        & Sim.  
        & 20k+ 
        & High 
        & BC 
        & 2B \\

        InternVLA-M1 \cite{chen2025internvla} 
        & Sim.  
        & \textless 10k 
        & High 
        & BC 
        & 3B \\

        Being-H0 \cite{luo2025being} 
        & Hum. 
        & \textless 10k 
        & Mixed 
        & Aln. + BC 
        & 14B \\

        InternVLA-A1 \cite{cai2026internvla} 
        & Het.  
        & 10k+
        & High 
        & VF + BC 
        & 3B \\

        GR00T-N1.6 \cite{bjorck2025gr00t} 
        & Het. 
        & \textless 10k 
        & Mixed 
        & LA + BC 
        & 1B \\ 

        UniVLA \cite{bu2025univla} 
        & Het.  
        & \textless 10k 
        & Mixed 
        & LA + BC 
        & 7B \\ 
        \midrule
        LDA-1B 
        & Het.  
        & \textbf{30k+} 
        & Mixed 
        & UWM \cite{zhu2025unified}
        & 1B \\
        \bottomrule
    \end{tabular}
    \caption{\textbf{Comparison of Representative Robot Foundation Models.}
This table compares the proposed LDA with recent robot foundation models in terms of data source, data quantity, action quality, training paradigm, and the number of trainable model parameters (excluding frozen components).
Data source abbreviations are as follows: Tele.=teleoperation, Sim.=simulation, Hum.=human demonstration, and Het.=heterogeneous data.
Training paradigm abbreviations include: BC=behavior cloning, VF=visual foresight, Aln.=alignment, LA=latent action modeling, and UWM=unified world model.
Only embodied interaction data are considered, excluding internet-scale VQA data.}
    \label{tab:ablation}
\end{table}


\noindent \textbf{Robot Foundation Models.} Recent robot foundation models predominantly adopt the Behavior Cloning paradigm. As summarized in Table~I, representative approaches including $\pi_0$~\cite{black2024pi_0}, RDT~\cite{liu2024rdt}, and InternVLA~\cite{chen2025internvla} rely heavily on high-quality teleoperation or simulation data, which fundamentally constrains their scalability. Hybrid methods such as Being-H0~\cite{luo2025being} and UniVLA~\cite{bu2025univla} attempt to incorporate heterogeneous data with mixed quality; however, they largely depend on action alignment or auxiliary pretrained latent action models, limiting the effective data scale to around 6k hours of embodied data. In contrast, LDA-1B breaks this ceiling by adopting a unified world model formulation, enabling efficient ingestion of up to 30k hours of mixed-quality embodied data.

\noindent\textbf{Unified Video Action Models.}
Recent works have explored joint modeling of dynamics and policy for embodied decision making. Methods such as DyWA~\cite{lyu2025dywa}, FLARE~\cite{zheng2025flare}, DreamVLA~\cite{dreamvla25} and the WorldVLA series~\cite{cen2025worldvla,jiang2025rynnvla} demonstrate that co-training next-state prediction and policy learning can improve generalization in interactive environments.
To enrich dynamics modeling, UWM~\cite{zhu2025unified} and UVA~\cite{li2025unified} further propose optimizing multiple objectives jointly, including video generation, forward and inverse dynamics, and action prediction. Concurrent with our work, Motus~\cite{bi2025motus} adopts the UWM paradigm and integrates priors from pretrained VLM and video generation models.
Despite their promising results, these approaches typically operate directly in pixel space and do not explicitly consider the roles of data quality, scale, or heterogeneity during training, which limits their ability to fully exploit large-scale, mixed-quality interaction data for robust dynamics learning.

\noindent\textbf{Large-Scale Embodied Interaction Datasets.}
The progress in embodied AI relies on large-scale embodied datasets.
Many widely used datasets are collected via teleoperation on real robots~\cite{black2024pi_0,intelligence2025pi_,kim2025openvla,liao2025genie} or generated in simulation~\cite{deng2025graspvla,chen2025internvla}, providing high-quality action-labeled trajectories.
Beyond robot-collected data, recent works explore human-centric embodied datasets, such as egocentric recordings with hand actions~\cite{yang2025egovla,luo2025being,fu2025metis}.
While these datasets significantly expand data diversity, many are either not publicly released or provide limited action supervision, making them difficult to directly integrate with robot learning pipelines.
More broadly, existing embodied datasets are highly fragmented: some are closed-source, others are open but vary substantially in data formats, sensor configurations, action representations, and annotation quality.
This lack of standardization poses a major obstacle to large-scale data aggregation and unified training.
In contrast, our work introduces EI-30k, a large-scale embodied interaction dataset that unifies diverse data sources including robot and human trajectories from both real-world and simulated environments under consistent data formats and aligned action representations.


\section{Latent Dynamics Action Model}
\subsection{Preliminary: Unified World Models}  
Given the current observation $o_t$ (typically an RGB image), UWM~\cite{zhu2025unified} jointly models multiple conditional distributions over future observations $\boldsymbol{o}_{t+1:t+k}$ and action chunk $\boldsymbol{a}_{t+1:t+k}$, enabling unified learning of:  
\begin{enumerate} 
    \item Policy: $p(\boldsymbol{a}_{t+1:t+k}\mid\boldsymbol{o}_t)$  
    \item Forward Dynamics: $p(\boldsymbol{o}_{t+1:t+k}\mid\boldsymbol{o}_t,\boldsymbol{a}_{t+1:t+k})$  
    \item Inverse Dynamics: $p(\boldsymbol{a}_{t+1:t+k}\mid\boldsymbol{o}_{t:t+k})$  
    \item Visual Planning: $p(\boldsymbol{o}_{t+1:t+k}\mid\boldsymbol{o}_t)$  
\end{enumerate}
Concretely, UWM~\cite{zhu2025unified} instantiates this framework using a joint diffusion model that predicts noise for both actions and future observations:  
\[
(\epsilon_a^{\theta}, \epsilon_o^{\theta})
= s_{\theta}\!\left(
o,\,
a_{t_a},\,
o^{\prime}_{t_o},\,
t_a,\,
t_{o^{\prime}}
\right),
\]  
where $t_a$ and $t_o$ are independently sampled diffusion timesteps for actions and observations, and $\tilde{a}_{t_a}$, $\tilde{o}_{t_o}$ denote their corresponding noisy inputs. The model is trained with a standard DDPM~\cite{ho2020denoising} objective, jointly denoising future actions and observations conditioned on $o_t$.  
We further extend this formulation by introducing language $\ell$ conditioning through a VLM, enabling instruction-guided action and observation prediction.

\subsection{Universal Data Ingestion via Multi-task Co-training}
We adopt a \emph{universal data ingestion} regime to jointly train the unified objectives described above, allowing heterogeneous embodied data to contribute according to their supervision quality.
Specifically, high-quality robot and human demonstrations are co-trained with all objectives, supporting both action policy learning and dynamics modeling.
Lower-quality trajectories, which may contain suboptimal or noisy actions, are used exclusively for dynamics and visual forecasting, where accurate action optimality is not required.
In addition, we leverage large-scale human manipulation videos without action annotations to train the visual forecasting objective, providing supervision for instruction-conditioned future state prediction.
This role-aware data usage prevents overfitting to expert-only behaviors and enables scalable learning of transferable dynamics and action representations.

To implement differentiated objectives within a single diffusion model, we introduce four learnable \emph{task embeddings} and two learnable \emph{register tokens}.
Each task embedding corresponds to a specific training objective (policy, forward dynamics, inverse dynamics, or visual forecasting) and is added to the diffusion timestep embedding $f_t$ to condition the denoising process.
The learnable register tokens, one for action and one for visual state, serve as placeholders for modalities that are absent in a given task.
For example, during policy training, the model receives noisy action tokens along with a visual register token representing the unobserved future state; in contrast, visual forecasting uses noisy future visual tokens with an action register token.
This design enables a unified architecture to flexibly support different input-output structures without modifying the network topology.
Overall, the model predicts a denoising vector field $v_a^\theta$ under different task conditions and is trained using a flow-matching objective:
\begin{equation}
\begin{aligned}
l_{\mathrm{action}}^\theta
&= \mathbb{E}_{\substack{
(\boldsymbol{o}_{t:t+k}, \boldsymbol{a}_{t+1:t+k}, \ell)\sim\mathcal{D} \\
\tau_a \sim \mathcal{U}(0, T_\tau) \\
\epsilon_a \sim \mathcal{N}(\boldsymbol{0}, \boldsymbol{I})
}}
\left\|
v_a^\theta - (\epsilon_a - \boldsymbol{a}_{t+1:t+k})
\right\|_2^2, \\
l_{\mathrm{obs}}^\theta
&= \mathbb{E}_{\substack{
(\boldsymbol{o}_{t:t+k}, \boldsymbol{a}_{t+1:t+k}, \ell)\sim\mathcal{D} \\
\tau_o \sim \mathcal{U}(0, T_\tau) \\
\epsilon_o \sim \mathcal{N}(\boldsymbol{0}, \boldsymbol{I})
}}
\left\|
v_o^\theta - (\epsilon_o - \boldsymbol{o}_{t+1:t+k})
\right\|_2^2, \\
l^\theta
&= l_{\mathrm{action}}^\theta + l_{\mathrm{obs}}^\theta .
\end{aligned}
\end{equation}

During training, action and visual losses are selectively activated according to the task specification, allowing heterogeneous data to contribute under appropriate supervision.
At inference time, the same model can be flexibly invoked for different objectives by specifying the task embedding and corresponding inputs.

\subsection{Representation of Predictive Targets}

We represent predictive targets, future visual states and actions, in a unified format to maximize knowledge sharing across heterogeneous datasets.  
For visual prediction, we adopt latent features extracted from a pretrained DINO~\cite{simeoni2025dinov3} encoder, rather than VAE-based pixel-space representations. DINO latents encode high-level semantic and spatial structure while suppressing background noise and low-level visual variations, which facilitates learning scene dynamics that generalize across diverse environments and object configurations.

For actions, we define a unified hand-centric action space based on end-effector motion, consisting of delta wrist poses and finger configurations. For parallel-jaw grippers, the finger state is represented by a single degree-of-freedom gripper width, while for multi-finger dexterous hands, finger configurations are described using keypoints expressed in the wrist coordinate frame. This design enables consistent action modeling across different embodiments and manipulation platforms.

To model temporal dynamics, visual states and actions are organized as two synchronized temporal streams with different sampling rates. Visual observations are sampled at 3 Hz, a lower frequency than actions, 10 Hz. This reduces redundant computation from highly correlated consecutive frames while preserving fine-grained action dynamics, allowing the model to maintain coherent temporal alignment between fast-varying control signals and slower-evolving visual states.

\subsection{Architecture: MM-DiT}

We adopt a Multimodal Diffusion Transformer (MM-DiT) to jointly denoise action chunks and predict future visual features within a unified diffusion framework (Fig.~2).
The model operates on heterogeneous tokens while sharing a common Transformer backbone.
Conditioning inputs include the current observation, language instruction, diffusion timestep, and task specification.
Observations and language are encoded by a pretrained VLM into conditioning tokens.
The diffusion timestep is encoded using a sinusoidal embedding, and task information is represented by a learned task embedding.
All conditioning signals are injected into each Transformer block via adaptive layer normalization (AdaLN~\cite{Peebles2022DiT}). 

Actions are organized as fixed-length chunks and corrupted with Gaussian noise.
Future visual features (DINO~\cite{simeoni2025dinov3} features) are noised in parallel.
Both modalities are projected into token embeddings through modality-specific linear layers and processed jointly by MM-DiT.
Each MM-DiT block applies multimodal self-attention over concatenated action and visual tokens, enabling cross-modal interaction.
Modality-specific QKV projections and FFNs are retained to preserve inductive biases, while attention is shared across modalities.
Language tokens are incorporated via cross-attention to provide high-level semantic guidance.
Finally, modality-specific output heads predict denoised action sequences and future visual features.

\subsection{Pre-training and Post-training}

\noindent\textbf{Pre-training Configurations.}
Our model is trained on a server cluster equipped with 48 NVIDIA H800 GPUs. The training process contains 400k iterations, resulting in a total computational cost of 4,608 GPU hours. To preserve the generalization capability and visual representation quality of the pre-trained foundation models, we keep the parameters of the VLM~\cite{yang2025qwen3} and the DINO~\cite{simeoni2025dinov3} encoder frozen throughout the pre-training process, updating the MM-DiT and action encoder/decoder. This design ensures that the model can learn from new data without degrading the core abilities of the base models in cross-modal understanding and fine-grained visual feature extraction.

\noindent\textbf{Data-Efficient Fine-tuning.} To adapt the model to target embodiments and tasks for real-world deployment, we introduce a lightweight post-training stage. This stage follows the same data regime as pretraining and effectively leverages naturally collected teleoperation data of mixed quality, without requiring expert-level demonstrations. Compared to prior fine-tuning pipelines that rely on carefully curated expert datasets, our method directly utilizes unfiltered teleoperation data, substantially improving data efficiency and reducing the cost of data collection and annotation, thereby facilitating practical deployment.

\begin{figure*}[htbp]
    \centering
    \includegraphics[width=0.95\linewidth]{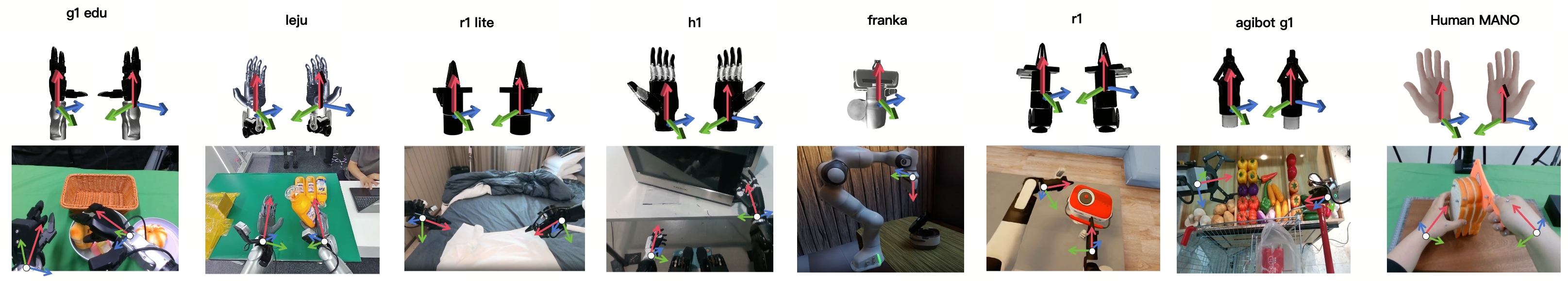}
    \caption{\textbf{Aligned End Effector Coordinate Systems.} We manually align coordinate frames across diverse robot and human embodiments to ensure consistency. This shared representation enables joint learning from heterogeneous interaction data.}
    \vspace{-2mm}
	\label{fig:eef}
\end{figure*}
\begin{figure}[t]
    \centering
    \includegraphics[width=\linewidth]{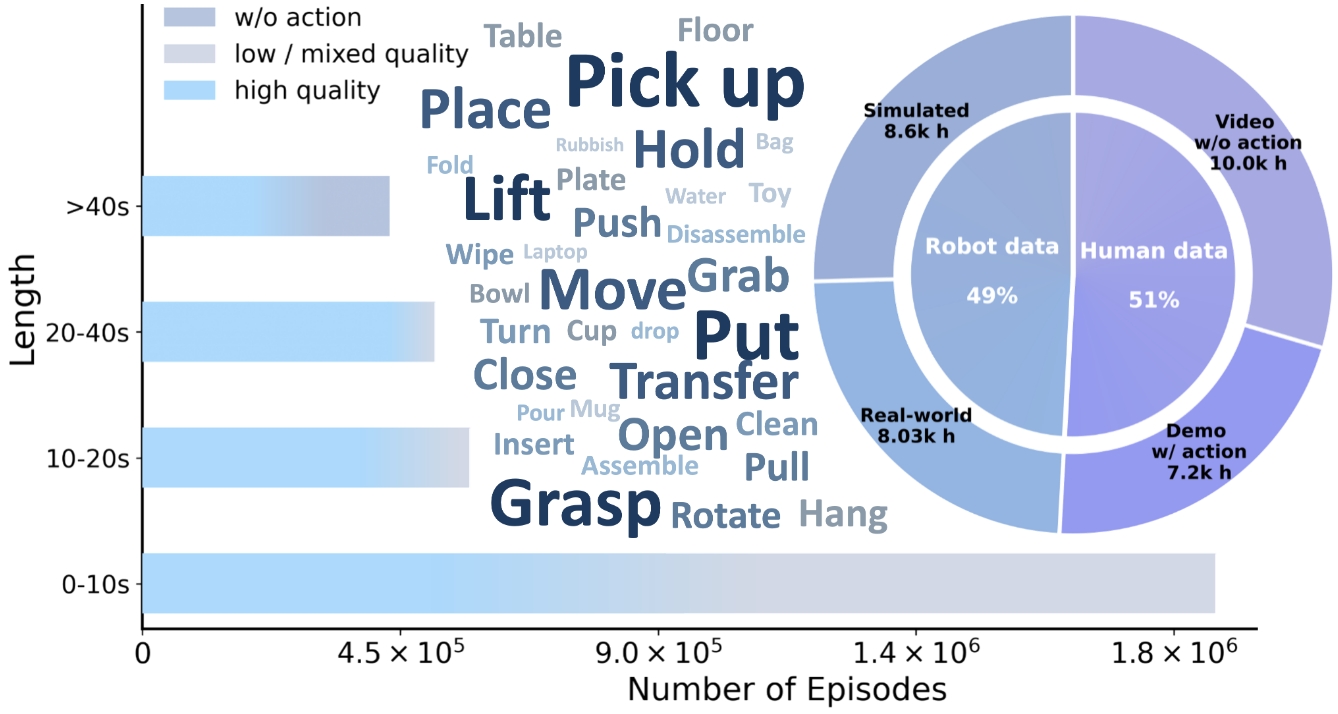}
    \caption{\textbf{Statistics of EI-30k.} The dataset contains more than 30k hours of diverse human and robot interaction data (right). It spans varying episode lengths (left) and a rich set of manipulation tasks (center).}
    \vspace{-5mm}
	\label{fig:EI30k}
\end{figure}

\section{Embodied Interaction Dataset (EI-30k)}

We introduce the Embodied Interaction Dataset (EI-30k), a large-scale collection of embodied interaction trajectories totaling over 30k hours.
It consists of 8.03k hours of real-world robot data, 8.6k hours of simulated robot data, 7.2k hours of human demonstrations with actions, and 10k hours of actionless human videos.  
All subdatasets are annotated with explicit quality labels, enabling systematic analysis across different fidelity levels and supporting quality-aware learning.

\noindent \textbf{Data Unification.}
EI-30k consolidates datasets from heterogeneous platforms and tasks, which vary in storage formats, sensor modalities, and annotations.
All data are converted into the LeRobot format, providing a unified representation of observations, actions, and language.  
This standardization facilitates plug-and-play training, flexible data composition, and seamless integration of additional annotations, while greatly reducing engineering overhead for handling diverse sources.

\noindent \textbf{Aligned Action Representation.}
To support consistent modeling of physical interactions across embodiments, all available action annotations are expressed as hand-centric motion in a shared coordinate frame (Fig. \ref{fig:eef}).  
For robots, this includes the 6-DoF end-effector pose plus gripper width or dexterous hand joints. For humans, the 6-DoF wrist pose and full MANO~\cite{MANO:SIGGRAPHASIA:2017} hand parameters are recorded.  
Camera extrinsics are retained to decouple hand motion from egocentric head motion. All coordinate frames are manually aligned to ensure geometric consistency across datasets, enabling joint learning from both human and robot trajectories.

\noindent \textbf{Quality Annotation and Cleaning.}
EI-30k applies systematic cleaning and quality-aware annotation.
Language annotations are normalized using a vision-language model to ensure semantic consistency. Motion segments without meaningful hand-object interaction are removed, e.g., head-only or idle segments in egocentric videos.  
Each trajectory is assigned a quality label based on action accuracy, and annotation completeness.  
Unlike aggressive filtering, low-quality trajectories are preserved, allowing downstream models to exploit the full spectrum of data through quality-aware training.

\section{Experiments}

\subsection{Simulation Experiments}

\noindent\textbf{Benchmark and Baselines.}
We evaluate our method on RoboCasa-GR1~\cite{robocasa2024}, a simulated kitchen benchmark featuring 24 tabletop rearrangement and articulated-object manipulation tasks with the GR-1 humanoid robot and Fourier dexterous hands.
The benchmark provides challenging and realistic settings that require high-DoF dexterous manipulation from egocentric RGB observations captured by a head-mounted camera.
Following the GR00T~\cite{bjorck2025gr00t} evaluation protocol, we finetune all models using 1,000 trajectories per task and evaluate each task with 51 trials, reporting average success rates.
We compare LDA against GR00T and its strong variants, as well as UWM~\cite{zhu2025unified}, under matched training paradigms and data.
To ensure a fair comparison in terms of model capacity and pretraining, we reproduce a strong GR00T baseline (denoted as GR00T-EI10k) with 1B parameters, pretrained on our curated EI-30k high-quality subset and using Qwen3-VL as the VLM encoder.

\setlength{\tabcolsep}{2.3pt}
\begin{table}[t]
    \centering
    {
    \begin{tabular}{lcccc}
    \toprule
    \rowcolor{gray!10}
        Model &Vis. Rep. & MMDiT  & VLM & Success Rate $\uparrow$ \\
    \midrule
        GR00T-N1.6~\cite{bjorck2025gr00t} & - & -  & Cosmos~\cite{azzolini2025cosmos, nvidia2025cosmosreason2}  & 47.6   \\
        StarVLA~\cite{community2026starvla, ye2026starvla}& - & -  & Qwen3-VL~\cite{yang2025qwen3}  & 47.8   \\
        GR00T-EI10k & - & - & Qwen3-VL  & 51.3\\ \midrule
        UWM-0.1B~\cite{zhu2025unified} & VAE & \xmark   & -  & 14.2   \\
        UWM-1B & VAE & \xmark   & Qwen3-VL  & 19.3   \\
        UWM(MM-DiT) &VAE & \cmark   & Qwen3-VL  & 20.0 \\
        LDA(DiT) &DINO & \xmark   & Qwen3-VL  & 48.9   \\
        LDA-0.5B &DINO & \cmark   & Qwen3-VL  & 50.7   \\
        \rowcolor{lightblue}
        LDA-1B &DINO & \cmark   & Qwen3-VL   & \textbf{55.4}   \\
    \bottomrule
    \end{tabular}}
    \caption{Results on RoboCasa-GR1~\cite{robocasa2024} and impact of state representation (VAE vs. DINO~\cite{simeoni2025dinov3}), model size, and the MM-DiT architecture on task success rates.}
    \label{tab:robocasa}
\end{table}

\begin{figure*}[t]
    \centering    \includegraphics[width=0.9\linewidth]{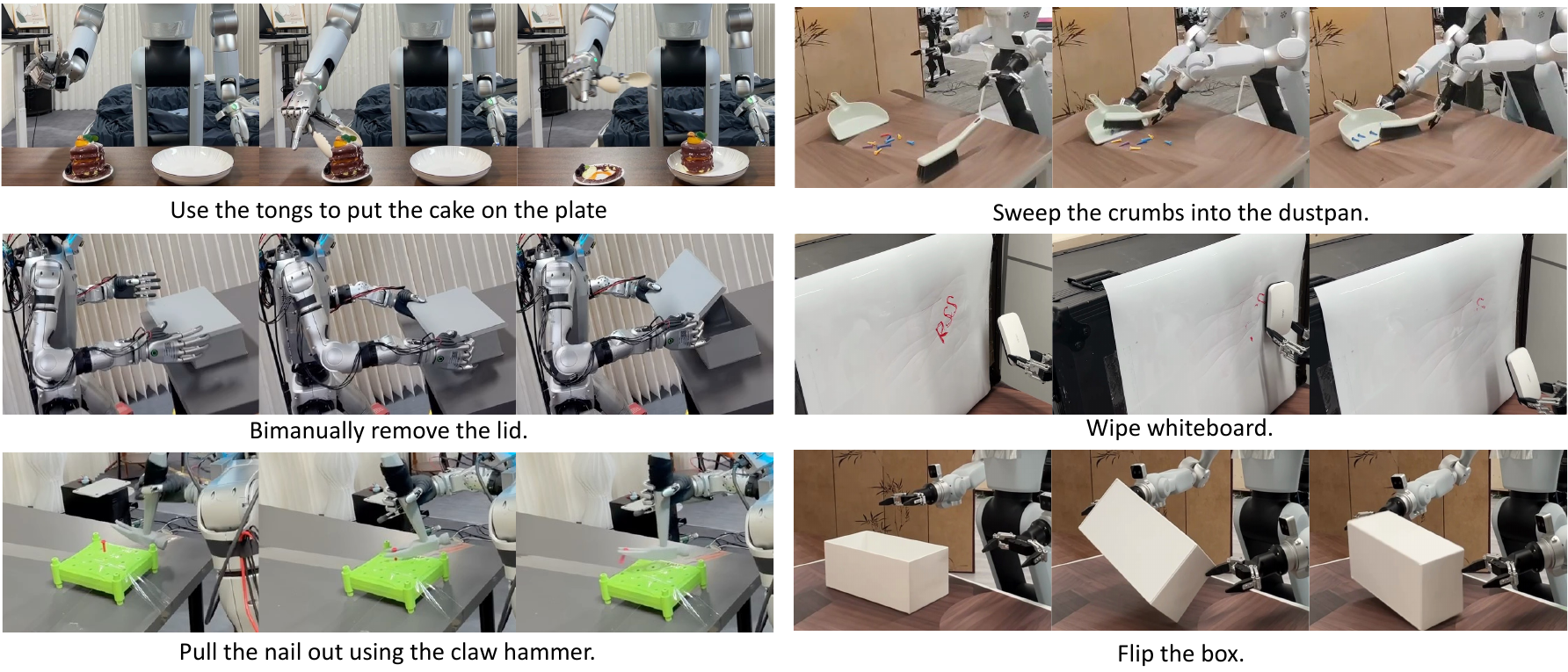}
    \vspace{-2mm}
    \caption{\textbf{Real-World Manipulation Demonstrations Across Multiple Robotic Platforms and End-Effectors.}
Galbot G1 equipped with a Sharpa dexterous hand (top-left), Unitree G1 with a BrainCo dexterous hand (middle and bottom-left), and Galbot G1 with a two-finger gripper (right).
}
	\label{fig:gallary}
\end{figure*}
\begin{figure*}[htbp]
    \centering    \includegraphics[width=0.95\linewidth]{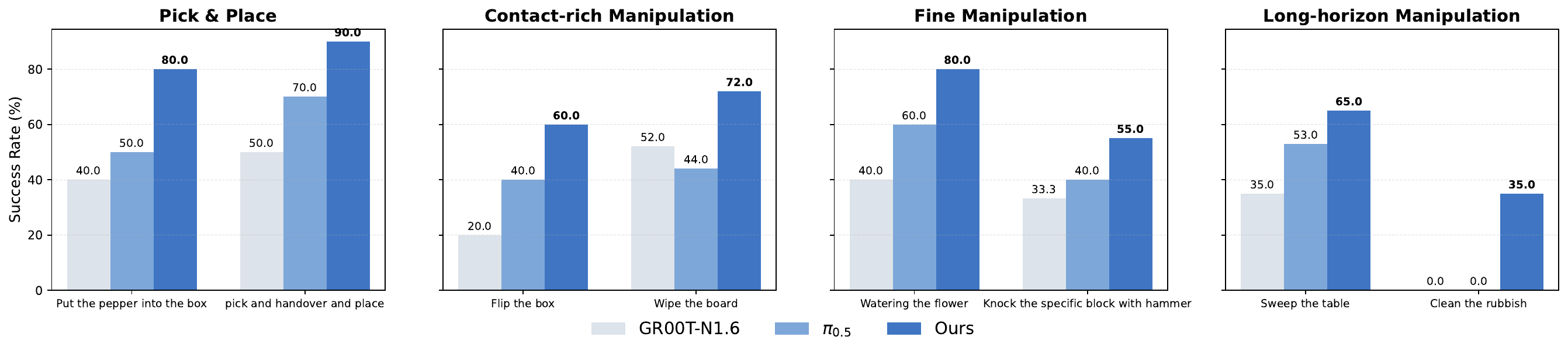}
    \vspace{-2mm}
    \caption{\textbf{Success Rate Comparison on Real-World Gripper Manipulation Tasks.}
All models are few-shot fine-tuned on Galbot and evaluated on eight tasks spanning Pick \& Place, Contact-rich, Fine, and Long-horizon manipulation. LDA consistently outperforms GR00T-N1.6~\cite{bjorck2025gr00t} and $\pi_{0.5}$~\cite{intelligence2025pi_}.}
	\label{fig:galbot_results}
    \vspace{-5mm}
\end{figure*}

\noindent\textbf{Comparison with Baselines.}
As shown in Table~\ref{tab:robocasa}, the original GR00T-N1.6~\cite{bjorck2025gr00t} with 3B parameters achieves a success rate of 47.6\%.
When pretrained on our curated EI-30k dataset, the reproduced GR00T-EI10k with 1B parameters shows a clear improvement, reaching 51.3\%, highlighting the impact of high-quality embodied data.
Under the same parameter budget, LDA further improves the success rate to 55.4\%.
These results indicate that, beyond data quality and parameter scaling, jointly learning actions and dynamics within a unified model provides additional gains when pretrained on mixed-quality data.

\noindent\textbf{Ablation Study.}
We further analyze key design choices under identical training data and optimization settings.
UWM~\cite{zhu2025unified}, despite jointly predicting actions and dynamics, achieves only 14.2\% success due to limited model capacity and the use of entangled VAE latent representations.
Scaling UWM to 1B parameters or replacing its DiT backbone with our MM-DiT yields only marginal improvements (19.3\% and 20.0\%, respectively), suggesting that architectural constraints fundamentally limit its performance.
In contrast, replacing pixel-space VAE latents with DINO~\cite{simeoni2025dinov3} representations leads to a substantial performance gain (20.0\% $\rightarrow$ 55.4\%), highlighting the importance of semantically structured latent spaces for effective scaling.
Finally, removing the proposed MM-DiT architecture or reducing the model size to 0.5B parameters results in performance drops of 6.5\% and 4.7\%, respectively, confirming the effectiveness of the multi-expert design and its favorable scaling behavior.

\subsection{Real-world Experiments}
\label{sec:experiments}

To validate the scalability and robustness of LDA-1B, we conduct extensive real-world experiments focusing on few-shot adaptation to new embodiments, dexterous manipulation, and data efficiency under mixed-quality supervision.

\noindent \textbf{Real-World Robot and Task Setup.} 
We evaluate our method on two humanoid platforms: Galbot G1 and Unitree G1.
Galbot G1 is equipped with either a two-finger gripper or 22-DoF Sharpa dexterous hands, while Unitree G1 uses 10-DoF BrainCo hands.
Across all configurations, the policy receives only egocentric RGB observations from a head-mounted camera.
We evaluate four categories of manipulation tasks under the gripper setting, \emph{Pick and Place}, \emph{Contact-rich Manipulation}, \emph{Fine Manipulation}, and \emph{Long-horizon Manipulation}, covering diverse contact dynamics and temporal horizons.
Representative tasks include \textit{Beat Block, Flip Box, Handover, Pick-and-Place (Pepper), Sweep Table, Clean Rubbish, Water Flower,} and \textit{Wipe Board}.
Dexterous manipulation further includes tool-use tasks such as pulling a nail with a hammer and flipping bread with a spatula, which require precise force control and coordinated finger motion.
Qualitative demonstrations are shown in Fig.~\ref{fig:gallary}.
For each task, we collect 100 teleoperated trajectories without enforcing expert-level execution.
As a result, the dataset naturally exhibits mixed quality: approximately 50--80\% of trajectories correspond to expert behavior, while the remainder contain suboptimal actions such as pauses, retries, or inefficient motion patterns.

\noindent \textbf{Baselines and Fine-tuning Protocol.}
We compare LDA-1B against two strong baselines, $\pi_{0.5}$~\cite{intelligence2025pi_} and GR00T~\cite{bjorck2025gr00t}.
To ensure stable and competitive performance, baseline models are finetuned exclusively on the filtered expert subset.
In contrast, LDA-1B leverages all collected trajectories and learns directly from the full mixed-quality distribution via our Universal Embodied Data Ingestion mechanism.

\begin{figure}[t]
    \centering
    \includegraphics[width=0.95\linewidth]{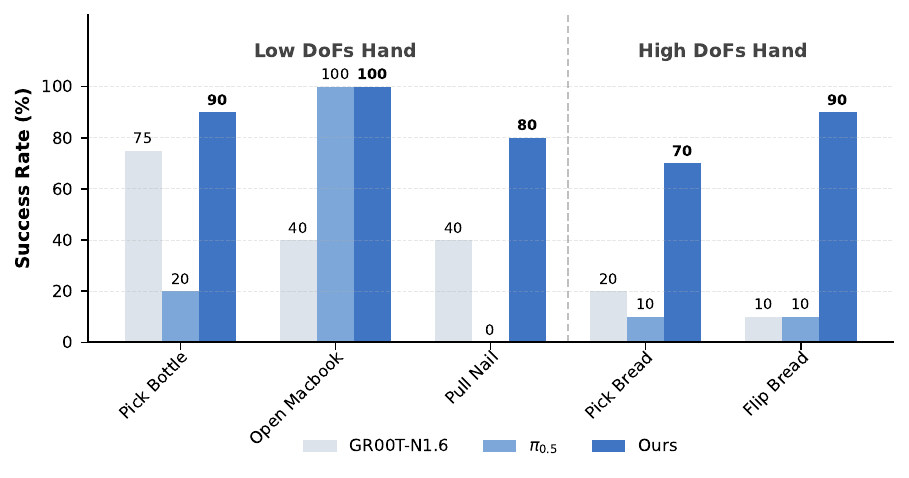}
    \caption{\textbf{Success Rate Comparison on Real-World Dexterous Manipulation Tasks.} We evaluate the real-world performance of our model against baselines (GR00T-N1.6 and $\pi_{0.5}$) on 3 low-DoF hand (BrainCo) tasks and 2 high-DoF hand (Sharpa) tasks. Ours (dark blue) consistently outperforms baselines, especially on fine dexterous tasks (pulling nails) and high-DoF tasks.}
    \vspace{-2mm}
	\label{fig:dexterous}
\end{figure}
 
\begin{figure*}[t]
    \centering
    \includegraphics[width=0.8\linewidth]{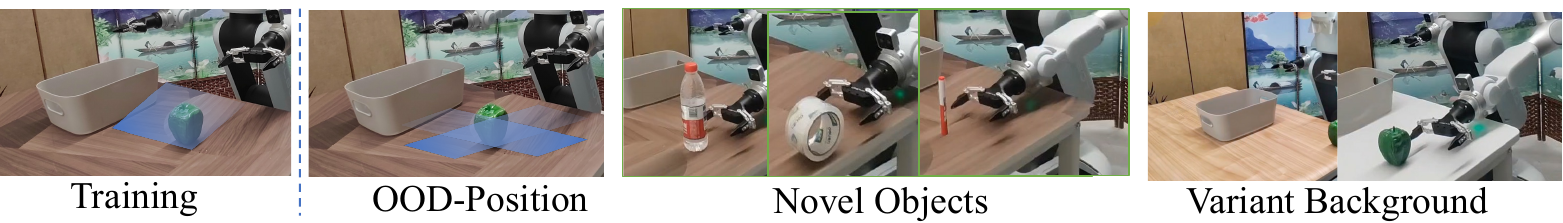}
    \caption{Generalization evaluation setup on Pick and Place task}
    \vspace{-2mm}
	\label{fig:OOD}
\end{figure*}
\begin{table}[t]
    \centering
    \label{tab:comparison_narrow}
    
    \setlength{\tabcolsep}{2pt}
    
    {%
        \begin{tabular}{lcccccc}
            \toprule
            \multirow{2}{*}{Method} & \multicolumn{3}{c}{Pick \& Place} \\
            \cmidrule(lr){2-4} 
            \rowcolor{gray!10}
             & Object & Background & OOD Pos.  \\
            \midrule
            $\pi_{0.5}$   & 26.7 & 20.0 & 6.7  \\
            GR00T & 40.0 & 40.0 & 20.0  \\
            \rowcolor{lightblue}
            Ours  & \textbf{60.0} & \textbf{60.0} & \textbf{40.0} \\
            \bottomrule
        \end{tabular}%
    }
    \caption{\textbf{Robust Generalization under visual and spatial perturbations.} LDA-1B achieves 60.0\% success on unseen objects and backgrounds, and 40.0\% under OOD positions, demonstrating effective focus on task-critical affordances over visual noise through latent dynamics pretraining.}
    \label{tab:generalization}
\end{table}

\noindent \textbf{Results on Gripper Manipulation.}
We first evaluate few-shot adaptation by deploying LDA-1B on the Galbot G1, which is excluded from our EI-30k pretraining dataset.
As shown in Fig.~\ref{fig:galbot_results}, LDA-1B consistently outperforms all baselines across task categories.
On simple pick-and-place tasks, LDA-1B achieves success rates of 80\%--90\%, indicating effective few-shot adaptation to a new robot embodiment.
The performance gap widens substantially in contact-rich and long-horizon scenarios.
For instance, the \textit{Clean the Rubbish} task requires coordinated dual-arm manipulation, tool usage (dustpan), and sequential object transfer into a trash bin, where errors can easily accumulate over time.
In this setting, LDA-1B achieves a 35\% success rate, while both GR00T and $\pi_{0.5}$ fail entirely (0\%).
This result suggests that latent dynamics modeling enables LDA to better anticipate action-induced state transitions, maintain temporal consistency, and recover from intermediate failures in extended manipulation sequences.

\noindent \textbf{Results on Dexterous Manipulation.}
We further evaluate LDA-1B on both low-DoF and high-DoF dexterous manipulation tasks, as reported in Fig.~\ref{fig:dexterous}.
On low-DoF tasks such as \textit{Pull Nail}, which requires precise motion direction and stable contact maintenance between the hammer and the nail, LDA-1B achieves 80\% success, reliably localizing targets and adjusting sensitive actions, whereas $\pi_{0.5}$ largely fails.
On high-DoF tasks such as \textit{Flip Bread}, which involve high-dimensional control, continuous contact, and coordinated wrist motion, LDA-1B attains 90\% success, while $\pi_{0.5}$ reaches only 10\%.
These results demonstrate that pretraining on large-scale human data provides strong latent priors for dexterous control, enabling precise finger coordination and object reorientation with limited robot data.
In contrast, baseline policies struggle to generalize as action dimensionality and contact complexity increase.

\begin{figure*}[t]
    \centering
    \includegraphics[width=0.95\linewidth]{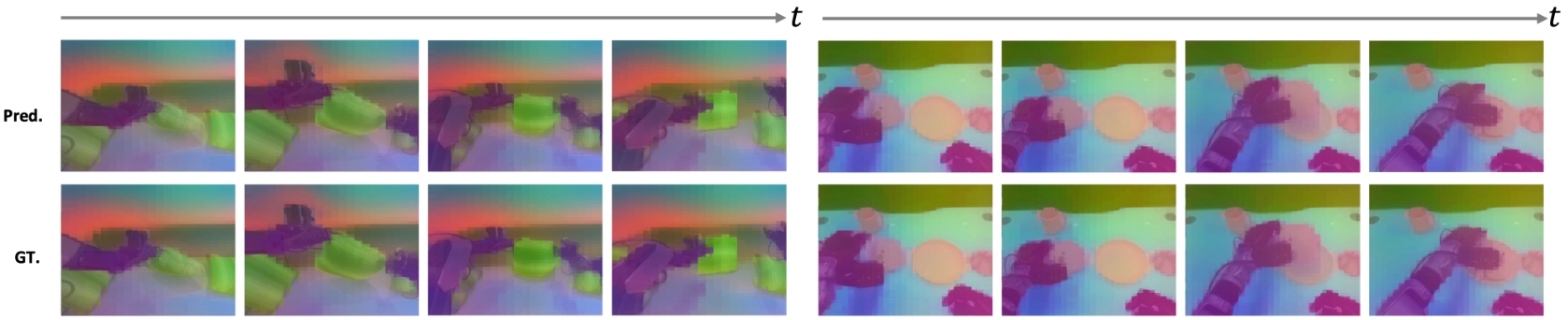}
    \caption{\textbf{Visualization of latent forward dynamics.} Our model generates accurate future visual representations (top) aligned with ground truth (bottom) across time steps, capturing semantic object structure and motion dynamics.}
    \vspace{-5mm}
	\label{fig:dino}
\end{figure*}

\begin{figure}[t]
    \centering
    \includegraphics[width=0.85\linewidth]{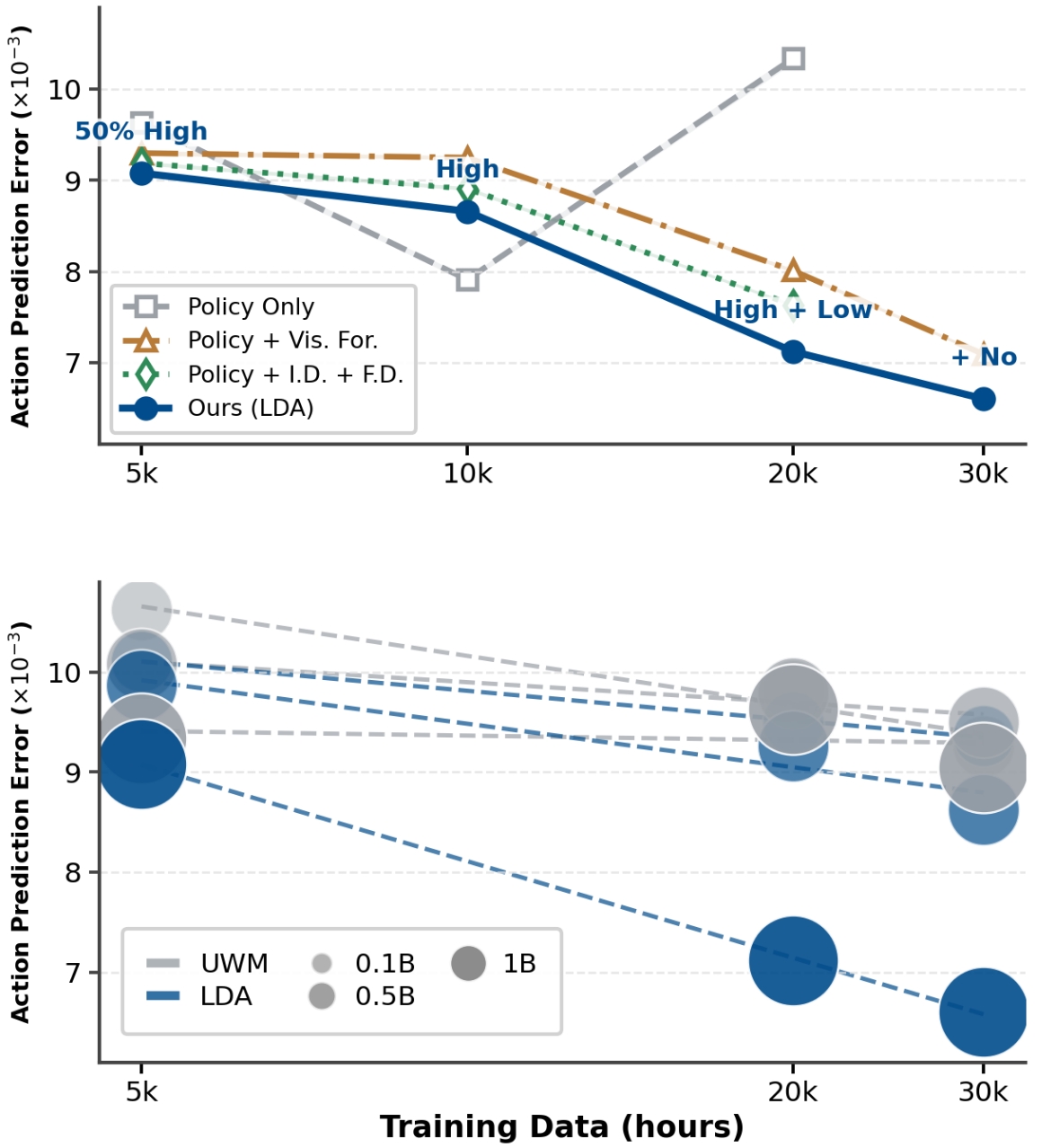}
    \caption{Scaling Analysis of LDA, evaluated by action prediction error on an unseen test set. Top: Action prediction error decreases to 6.6 with 30k hours of training data, demonstrating effective utilization of diverse data sources. Bottom: LDA consistently outperforms UWM across model sizes (0.1B$\rightarrow$1B) with increasing training data.}
    \vspace{-3mm}
    \label{fig:scaling_analysis}
\end{figure}

\noindent \textbf{Generalization Ability.}
To evaluate the generalization of our policy, we test pick and place task under three conditions: novel objects, unseen backgrounds, and out-of-distribution (OOD) starting position, shown in Fig.~\ref{fig:OOD}.
As summarized in Table~\ref{tab:generalization}, our model maintains high success rates despite visual and spatial perturbations. The large-scale latent dynamics pretraining allows the model to ignore visual distractors (background changes) while focusing on relevant object affordances, demonstrating strong generalization relative to baselines.

\noindent \textbf{Data-Efficient Fine-tuning.}
\begin{table}[t]
    \centering
    \label{tab:finetune_tasks}
    
    \setlength{\tabcolsep}{3pt}
    
    {%
        \begin{tabular}{lcccc}
            \toprule
            \multirow{2}{*}{Method} & \multicolumn{2}{c}{Place the pen into the box} & \multicolumn{2}{c}{Bimanually remove the lid} \\
            \cmidrule(lr){2-3} \cmidrule(lr){4-5}
             & 63 High & 63 High + 37 Low & 66 High & 66 High + 34 Low \\
            \midrule
            $\pi_{0.5}$ & 60 & 40 (20$\downarrow$) & 50 & 40 (10$\downarrow$) \\
            \rowcolor{lightblue} 
            Ours & \textbf{70} & \textbf{80 (10$\uparrow$)} & \textbf{50} & \textbf{60 (10 $\uparrow$)} \\
            \bottomrule
        \end{tabular}%
    }
    \caption{\textbf{Data-efficient mixed-quality fine-tuning.} LDA-1B improves success rates by +10\% on both tasks when incorporating low-quality trajectories, while $\pi_{0.5}$ degrades significantly, demonstrating effective utilization of noisy data for enhanced generalization.}
    \label{tab:data_efficiency}
    
\end{table}

We analyze the value of mixed-quality data ingestion during the fine-tuning stage by post-training on two splits: (1) High-Quality Only (expert data), and (2) High + Low Quality (all 100 trajectories).
As shown in Table~\ref{tab:data_efficiency}, while baseline models degrade when low-quality data is added, LDA-1B effectively leverages these noisy trajectories, boosting performance by 10 percentage points, substantially improving data efficiency and reducing the cost of data collection and annotation for practical deployment.


\subsection{Analysis of Design Choices for Model Scaling}
\label{subsec:dynamics_analysis}
To analyze the scaling behavior of LDA, we systematically vary model capacity, data composition, and training objectives.
All models are evaluated on an unseen test set sampled from a held-out subset of \textit{Agibot World}~\cite{bu2025agibot_iros}.
We report the action prediction L1 error as the primary metric, which serves as a stable and reproducible proxy for real-world performance.
Fig.~\ref{fig:scaling_analysis} summarizes the results under four training configurations:
(i) \textit{Policy Only}, 
(ii) \textit{Policy + Visual Forecasting}, 
(iii) \textit{Policy with Forward and Inverse Dynamics}, and 
(iv) the full co-training framework (\textit{Ours}).
These experiments jointly reveal how LDA scales under heterogeneous supervision and increasing model capacity.

\noindent \textbf{Effectiveness of Universal Data Ingestion.}  
Effectively leveraging heterogeneous embodied data requires jointly scaling both data sources and training objectives.
As shown in Fig.~\ref{fig:scaling_analysis}, LDA achieves its best performance only when all supervision signals-policy learning, dynamics modeling, and visual forecasting-are optimized together.
When either the data scale or the training objectives are reduced, performance degrades noticeably.
Using only action-labeled trajectories with a \textit{Policy Only} objective (grey line), increasing the dataset size yields unstable behavior: while moderate scaling initially reduces error, incorporating lower-quality data leads to performance degradation.
Similarly, partial co-training variants that exclude either dynamics or visual forecasting objectives (green and brown lines) improve robustness but fail to fully exploit the available data.
In contrast, the full co-training framework (blue line) exhibits consistent improvement as additional heterogeneous data is introduced.
Notably, even after all action-labeled trajectories are exhausted, adding 10k actionless videos continues to reduce prediction error.
These results indicate that LDA can extract useful supervisory signals from low-quality data and non-action data through latent dynamics and visual forecasting, rather than treating such data as noise.
Overall, these results demonstrate that Universal Data Ingestion is most effective when heterogeneous data and co-training objectives are scaled together, enabling LDA to fully utilize mixed-quality supervision.

\noindent \textbf{Effectiveness of Latent Representation.}  
Although both LDA and UWM incorporate dynamics-related supervision, their scaling behaviors diverge substantially due to differences in the structure of their latent spaces.
As shown in Fig.~\ref{fig:scaling_analysis}, UWM quickly saturates as data scale and model capacity increase, with additional supervision yielding diminishing or even negative returns.
This indicates that simply increasing data or parameters is insufficient when the latent space cannot support compositional and causal reasoning.
This limitation stems from UWM's VAE-derived latent representation, which entangles appearance, geometry, and dynamics at a low-level feature granularity.
Such entanglement restricts the model’s ability to factorize action-induced state transitions and prevents effective reuse of heterogeneous supervision during scaling.
In contrast, LDA operates in a semantically structured latent space obtained from large-scale visual pretraining.
This representation preserves object-level semantics and spatial coherence, enabling dynamics learning to scale smoothly with increased model capacity, richer training objectives, and more diverse datasets.

\noindent \textbf{Effectiveness of Model Size Scaling.}  
Beyond data scale, LDA exhibits consistent and predictable improvements as model capacity increases.
As shown in Fig.~\ref{fig:scaling_analysis}, scaling the model from 0.1B to 0.5B and further to 1B parameters leads to monotonic reductions in action prediction error under the full co-training framework.
This indicates that LDA can effectively absorb additional capacity to model increasingly complex action-dynamics relationships when sufficient heterogeneous supervision is available. The results highlight a promising scaling paradigm in which model capacity, training objectives, and heterogeneous embodied data are jointly aligned, enabling reliable performance gains.

\subsection{Analysis of Multi-task Learning}
\noindent \textbf{Qualitative Analysis of Latent Forward Dynamics.}
Beyond quantitative prediction errors, we qualitatively examine the forward dynamics learned by LDA, visualized via PCA projections of DINO feature embeddings.
As shown in Fig.~\ref{fig:dino}, the model produces coherent future-state predictions that respect physical constraints such as object permanence, contact continuity, and motion consistency under the applied action.
Notably, the predicted dynamics focus on task-relevant objects while remaining invariant to visual distractors that do not influence the control loop.
This suggests that LDA learns a dynamics-aware latent world model, capturing how actions causally propagate through the scene rather than merely extrapolating visual appearance.

\begin{figure}[t]
    \centering
    \includegraphics[width=0.95\linewidth]{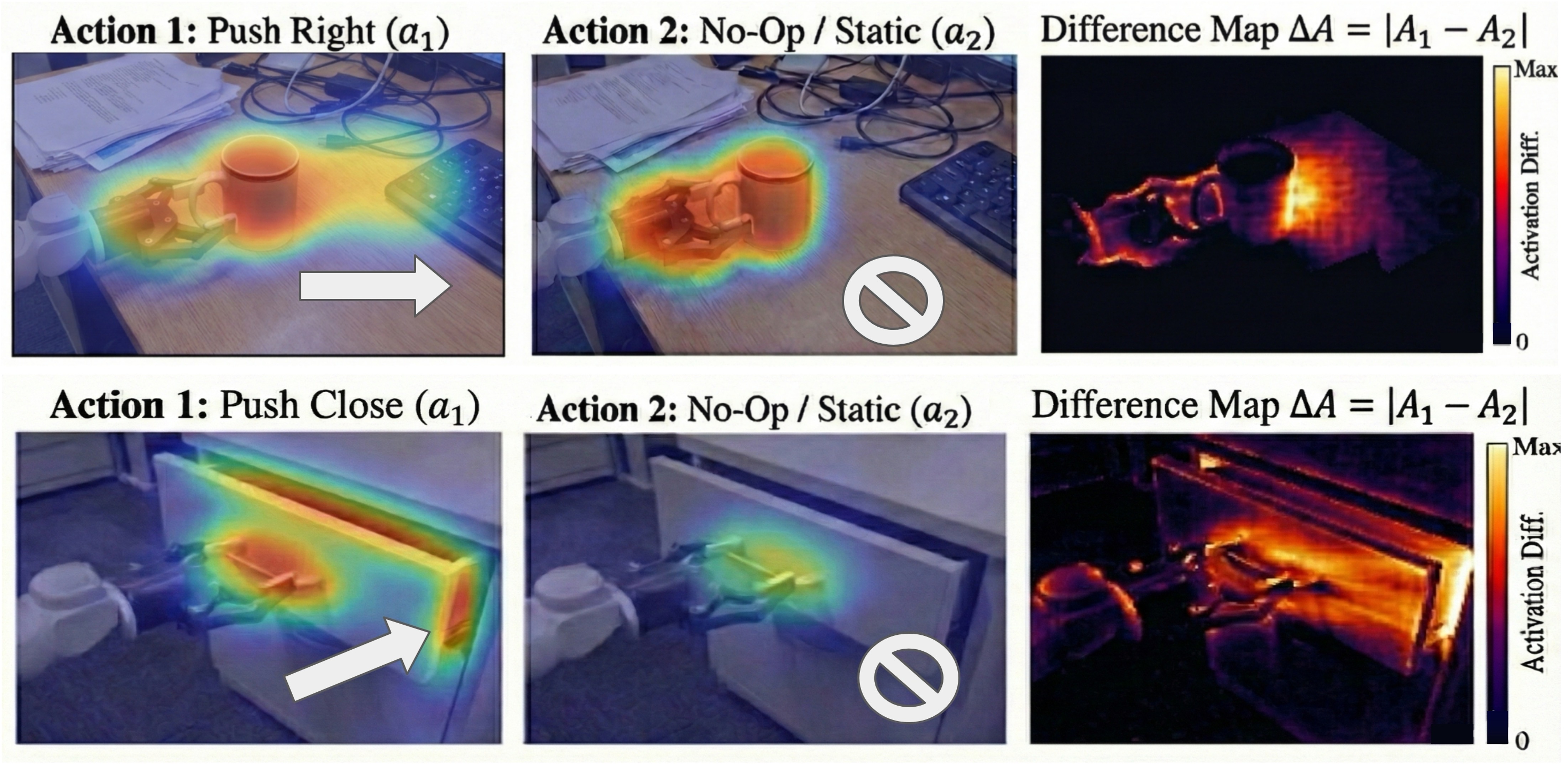}
    \caption{Attention Heatmap: ``Push Right'' (top) highlights the mug's leading edge and trajectory; ``Push Close'' (bottom) concentrates on the contact surface. The model attends exclusively to movable regions while ignoring irrelevant background clutter.}
    \vspace{-2mm}
	\label{fig:attn_diff}
\end{figure}

\noindent \textbf{Action-Conditioned Attention.} 
To interpret how LDA reasons about action-induced state transitions, we visualize attention maps conditioned on different action primitives.
As shown in Fig.~\ref{fig:attn_diff}, we compare the attention patterns induced by an active motion command ($a_1$) with those under a static \emph{No-Op} command ($a_2$), and compute their difference to reveal action-specific visual grounding.
Across tasks, LDA consistently attends to regions that are causally relevant to the commanded interaction.
In the \emph{Push Right} scenario, the attention difference highlights the leading edge of the mug and the anticipated motion direction, reflecting awareness of object displacement.
In the \emph{Push Close} task, attention concentrates on the drawer surface where contact and force application are expected.
Importantly, background clutter and visually salient but non-interactive regions are largely suppressed.
These results indicate that LDA conditions visual attention on the physical consequences of actions, selectively focusing on regions that drive state transitions rather than static appearance.

\begin{figure}[t]
    \centering
    \includegraphics[width=0.92\linewidth]{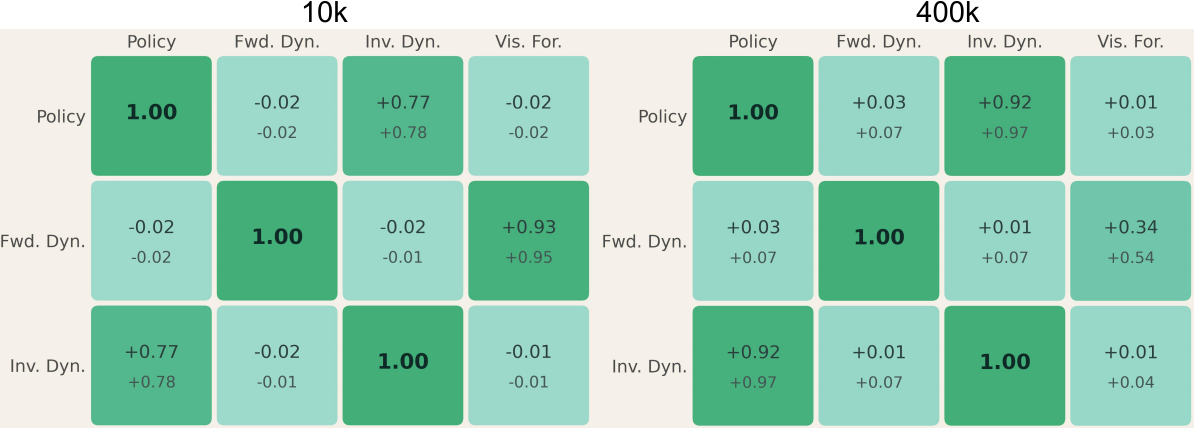}
    \caption{Gradient cosine similarities between objectives at 10k and 400k iterations.}
    \label{fig:action_gradient_similarity}
    \vspace{-3mm}
\end{figure}

{
\noindent \textbf{Gradient Similarity Analysis.} 
We measure gradient cosine similarities among training objectives of the action expert at 10k and 400k iterations. Fig.~\ref{fig:action_gradient_similarity} shows that objectives with matched targets are strongly aligned from early training and become more consistent over time: policy and inverse dynamics both predict actions, while video generation and forward dynamics both predict DINO features.
Objectives with different prediction targets show weak negative correlations at 10k iterations, but this early competition disappears by 400k iterations and turns into positive alignment.
This trend suggests that LDA gradually organizes heterogeneous objectives into compatible feature subspaces.
}


\IEEEpeerreviewmaketitle

\section{Conclusion, Limitations, and Future Directions}
\label{sec:conclusion}

We present \textbf{LDA-1B}, a robot foundation model that scales latent dynamics learning via universal embodied data ingestion. By assigning heterogeneous data distinct roles and leveraging over \textbf{30k hours} of human and robot trajectories in the EI-30k dataset, LDA-1B learns dynamics in a structured DINO latent space and employs a mixed-frequency multimodal diffusion transformer, enabling stable training at the \textbf{1B-parameter} scale. Experiments show strong performance across diverse manipulation and long-horizon tasks, as well as data-efficient fine-tuning on imperfect trajectories.
Limitations include the reliance on fixed DINO visual features and predominantly egocentric camera viewpoints, which may constrain generalization to new visual perspectives and multimodal signals. Future work includes jointly learning visual representations and latent dynamics, extending to richer sensory modalities, automatically optimizing data roles, and fostering broader community adoption of scalable, heterogeneous data-driven robot foundation models.

\section*{Acknowledgments}
{This work was partially supported by New Generation Artificial Intelligence-National Science and Technology Major Project (No. 2025ZD0122905). We thank Caowei Meng for teleoperation data, Haoran Liu and Jiayi Su for early exploration assistance, Yu-Wei Chao and Shengliang Deng for discussions, and Junkai Zhao for experimental equipment.}


\bibliographystyle{plainnat}
\bibliography{references}

\clearpage
\begin{appendices}

\section{Details of Model}
We employ Qwen3-VL-4B-Instruct~\cite{yang2025qwen3} as the joint language and vision encoder to extract high-level semantic representations. Visual observations are encoded using DINOv3-ViT-s~\cite{simeoni2025dinov3}. During pretraining, we freeze both the VLM and the DINOv3 image encoder to leverage the strong priors from the pretrained language and vision models while allowing the MM-DiT to be trained thoroughly on the downstream structure. In the subsequent finetuning stage, we unfreeze the VLM to enable end-to-end adaptation and further improve overall performance. 

Additionally, the MM-DiT is conditioned on a short history of two timesteps, comprising both past DINO-encoded observations and actions, to effectively capture temporal dynamics. Table~\ref{tab:model}  presents the detailed configurations of the model and the hyperparameters used during training.

\begin{table}[htbp]
    \centering
    \begin{tabular}{lc}
    \toprule
    \rowcolor{gray!10}
        \textbf{Parameter} & \textbf{Value} \\
    \midrule
        \textbf{Model}  \\
        VLM & Qwen3-VL~\cite{yang2025qwen3} \\
        Observation Encoder & DINOv3-ViT-s~\cite{simeoni2025dinov3} \\
        Hidden Size & 1536 \\
        Layers & 16 \\
        Attention Heads & 32 \\
        Image Shape & (224, 224, 3) \\
        Latent Image Shape & (14, 14, 384) \\
        Action Chunk & 16 \\
    \midrule
        \textbf{Training} \\
        Batch Size & 32 * 48 (pretraining) \\
        & 12 * 8 (fine-tuning) \\
        Learning Rate & $1e^{-4}$ \\
        Optimizer & AdamW \\
        Weight Decay & $1e^{-5}$ \\
        Betas & [0.9, 0.95] \\
        Epsilon & $1e^{-8}$ \\
        LR Schedule & cosine w/ min lr \\
        Min LR & $5e^{-7}$ \\
        
    \bottomrule
    \end{tabular}
    \caption{Model and Training configuration hyperparameters}
    \label{tab:model}
\end{table}

\begin{figure*}[t]
    \centering
    \includegraphics[width=0.90\linewidth]{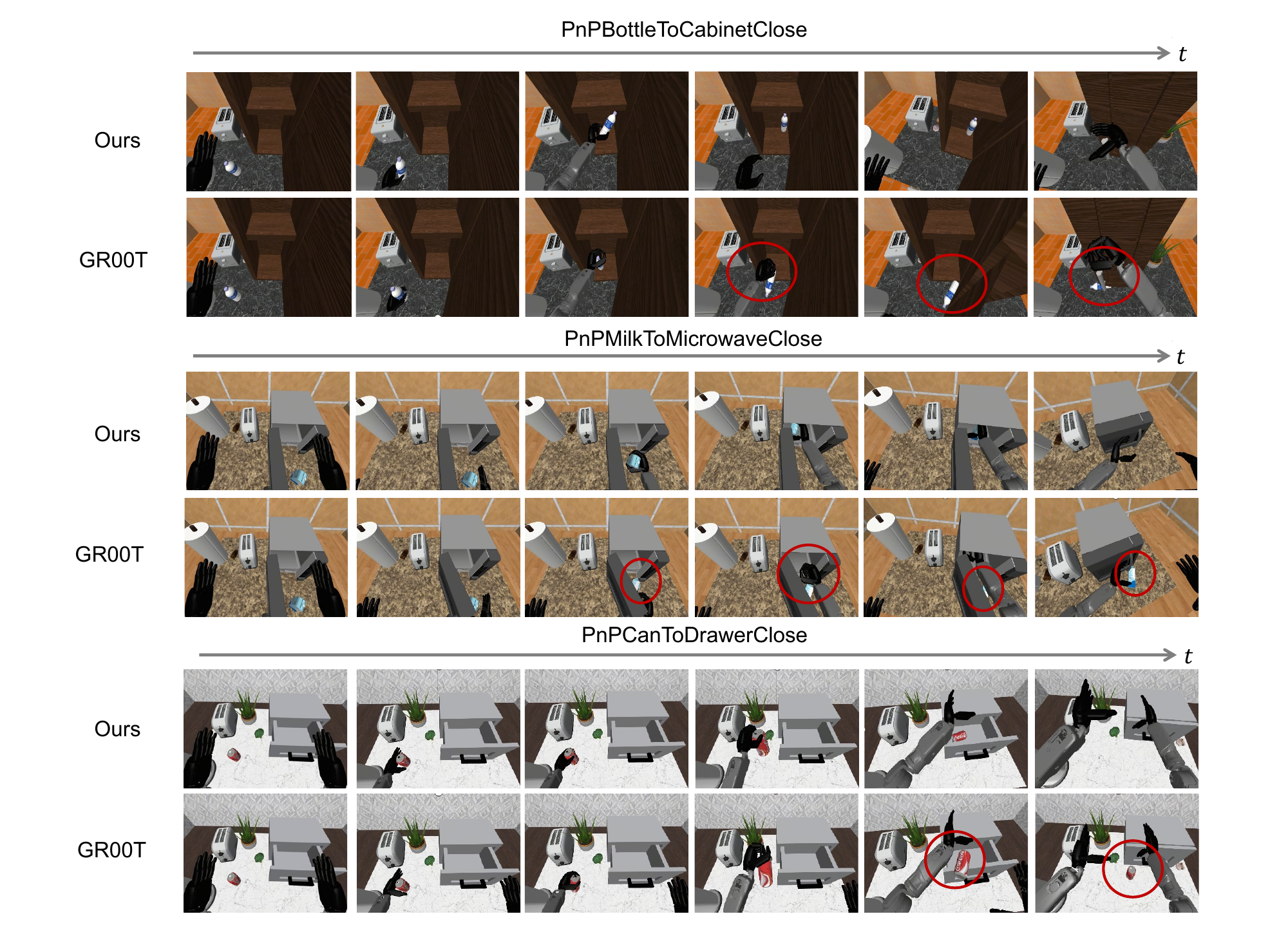}
    \caption{Qualitative comparison between our model and GR00T~\cite{bjorck2025gr00t} on RoboCasa-GR1~\cite{robocasa2024} manipulation tasks.
    Three representative tasks demonstrate our model's superior robustness in object grasping and placement accuracy. Critical failure modes of GR00T, including grasp slippage, misaligned object placement, and collision during manipulation, are highlighted with circles, while our model consistently achieves successful task completion.}
    \vspace{-2mm}
	\label{fig:robocasa_demo}
\end{figure*}
\begin{table*}[!ht]
    \centering
    \begin{tabular}{lcccccccc}
        \toprule
        model 
        & UWM & UWM-XL & UWM+MM-DiT 
        & GR00T & StarVLA & GR00T-EI10k 
        & LDA (DiT) & LDA \\ \midrule

        PnP Bottle To Cabinet Close
        & 27 & 41 & 49
        & 51.5 & 46 & \second{69}
        & \second{65} & \best{76} \\

        PnP Can To Drawer Close
        & 22 & 53 & 55
        & 13 & \best{80} & \second{61}
        & 59 & \second{71} \\

        PnP Cup To Drawer Close
        & 18 & 12 & 43
        & 8.5 & \best{54} & \second{47}
        & 40 & 41 \\

        PnP Milk To Microwave Close
        & 22 & 25 & 33
        & 14 & \second{48} & \best{75}
        & 47 & \second{52} \\

        PnP Potato To Microwave Close
        & 16 & 29 & 18
        & \best{41.5} & 28 & \second{41}
        & 39 & \second{41} \\

        PnP Wine To Cabinet Close
        & 31 & 24 & 25
        & 16.5 & 46 & \second{51}
        & 49 & \best{57} \\

        PnP Novel From Cuttingboard To Basket
        & 8 & 18 & 10
        & \second{58} & 48 & 43
        & 55 & \best{65} \\

        PnP Novel From Cuttingboard To Cardboardbox
        & 8 & 14 & 16
        & 46.5 & 40 & 39
        & \second{57} & \best{69} \\

        PnP Novel From Cuttingboard To Pan
        & 24 & 20 & 27
        & \second{68.5} & \second{68} & 67
        & 65 & \best{75} \\

        PnP Novel From Cuttingboard To Pot
        & 16 & 25 & 20
        & \best{65} & 52 & 53
        & \second{57} & \second{61} \\

        PnP Novel From Cuttingboard To Tieredbasket
        & 10 & 10 & 6
        & 46.5 & \best{56} & 29
        & 39 & \second{51} \\

        PnP Novel From Placemat To Basket
        & 8 & 16 & 14
        & \best{58.5} & 42 & 45
        & 37 & \second{53} \\

        PnP Novel From Placemat To Bowl
        & 12 & 10 & 14
        & \best{57.5} & 44 & \second{55}
        & 53 & \second{55} \\

        PnP Novel From Placemat To Plate
        & 10 & 12 & 10
        & \best{63} & 48 & \second{57}
        & 51 & \second{59} \\

        PnP Novel From Placemat To Tieredshelf
        & 2 & 2 & 2
        & \best{28.5} & 18 & \second{20}
        & 22 & \second{24} \\

        PnP Novel From Plate To Bowl
        & 12 & 8 & 14
        & \second{57} & \best{60} & 49
        & \second{57} & 53 \\

        PnP Novel From Plate To Cardboardbox
        & 2 & 10 & 8
        & 43.5 & \second{50} & \best{61}
        & 43 & 43 \\

        PnP Novel From Plate To Pan
        & 10 & 20 & 16
        & 51 & \second{54} & 51
        & 49 & \best{55} \\

        PnP Novel From Plate To Plate
        & 22 & 27 & 25
        & \best{78.7} & \second{70} & 67
        & 59 & 61 \\

        PnP Novel From Tray To Cardboardbox
        & 20 & 25 & 20
        & \second{51.5} & 38 & 49
        & \second{59} & \best{65} \\

        PnP Novel From Tray To Plate
        & 12 & 18 & 16
        & \best{71} & 56 & 57
        & \second{57} & \second{63} \\

        PnP Novel From Tray To Pot
        & 18 & 25 & 20
        & \best{64.5} & 50 & \second{63}
        & 53 & 55 \\

        PnP Novel From Tray To Tieredbasket
        & 6 & 16 & 16
        & \best{57} & 36 & \second{55}
        & 39 & 51 \\

        PnP Novel From Tray To Tieredshelf
        & 4 & 2 & 4
        & \second{31.5} & 16 & \second{31}
        & 22 & \best{33} \\

        \midrule
        Average
        & 14.3 & 19.3 & 20.0
        & 47.6 & 47.8 & \second{51.3}
        & 48.9 & \best{55.4} \\

        \bottomrule
    \end{tabular}
    \caption{Results on RoboCasa-GR1~\cite{robocasa2024} benchmark. UWM: UWM~\cite{zhu2025unified} with 140M parameters.
UWM-XL: UWM with 1B parameters, using Qwen3-VL~\cite{yang2025qwen3} as the joint encoder for language instructions and visual inputs.
UWM+MM-DiT: UWM-XL with its DiT~\cite{Peebles2022DiT} backbone replaced by our MM-DiT architecture.
StarVLA: GR00T~\cite{bjorck2025gr00t} equipped with Qwen3-VL as its System 2 module.
GR00T-EI10k: GR00T~\cite{bjorck2025gr00t} pretrained on our dataset and equipped with Qwen3-VL as its System 2 module.
LDA (DiT): Our LDA model with the MM-DiT replaced by a standard DiT. During finetuning on RoboCasa, the VLM is unfrozen to enable end-to-end adaptation.}
\label{tab:detail_robocasa} 
\end{table*}

\section{Detailed Results on the Simulation Benchmark}

\subsection{Evaluation Setup and Model Description.}
All methods are evaluated on the full set of 24 RoboCasa-GR1~\cite{robocasa2024} tasks, with 51 evaluation trials per task.
Unless otherwise specified, models are finetuned using 1,000 demonstrations per task and optimized under the same training paradigm to isolate architectural differences.

We summarize the evaluated models below:
\begin{itemize}
    \item \textbf{UWM}: A 140M-parameter Unified World Model~\cite{zhu2025unified}, serving as a lightweight baseline.
    \item \textbf{UWM-XL}: A 1B-parameter UWM variant equipped with Qwen3-VL~\cite{yang2025qwen3} for joint language-vision encoding.
    \item \textbf{UWM+MM-DiT}: UWM-XL with its DiT backbone replaced by our MM-DiT architecture.
    \item \textbf{GR00T-N1.6}~\cite{bjorck2025gr00t}: The original GR00T policy model without explicit dynamics modeling.
    \item \textbf{StarVLA}: A GR00T variant following StarVLA~\cite{community2026starvla, ye2026starvla}, replacing the original VLM with Qwen3-VL and trained from scratch on RoboCasa.
    \item \textbf{GR00T-EI10k}: A strong reproduced baseline pretrained on our EI-10k high-quality subset with Qwen3-VL, where VLM parameters are unfrozen during finetuning.
    \item \textbf{LDA (DiT)}: An ablated version of LDA replacing MM-DiT with a standard DiT backbone.
    \item \textbf{LDA-1B}: The full Latent Dynamics Action model with MM-DiT, designed to model action-induced state transitions in a structured latent space.
\end{itemize}

\subsection{Task-Level Results and Analysis.}
Table~\ref{tab:detail_robocasa} reports detailed per-task success rates.
LDA consistently outperforms GR00T across contact-rich and cluttered rearrangement tasks, with particularly large gains in scenarios requiring precise placement and closing actions, such as
PnP Bottle To Cabinet Close (76 \% vs. 51.5\%),
PnP Can To Drawer Close (71\% vs. 13\%),
and PnP Milk To Microwave Close (52\% vs. 14\%).

As illustrated in Fig.~\ref{fig:robocasa_demo}, GR00T frequently fails due to a lack of anticipation of post-action consequences.
For example, after placing an object inside a container, GR00T often retracts its arm along a trajectory that collides with the object, causing it to tip over.
In contrast, LDA anticipates such interactions and generates trajectories that preserve object stability throughout the entire manipulation sequence.

The largest improvements are observed in novel-object rearrangement tasks involving transfers across surfaces and containers
(e.g., \textit{Cuttingboard} $\rightarrow$ \textit{Basket}/\textit{Cardboardbox},
\textit{Placemat} $\rightarrow$ \textit{Plate}/\textit{Tieredshelf},
and \textit{Tray} $\rightarrow$ \textit{Cardboardbox}/\textit{Plate}/\textit{Pot}).
These tasks require adaptive contact handling and trajectory correction under clutter, where LDA shows clear advantages.
While GR00T remains competitive on a small subset of simple pick-and-place tasks with minimal environmental interaction, these cases are limited.
Overall, LDA's higher average success rate (55.4\% vs. 47.6\%) reflects a systematic advantage in complex and contact-rich manipulation scenarios rather than isolated gains.

\begin{figure*}[t]
    \centering
    \includegraphics[width=0.90\linewidth]{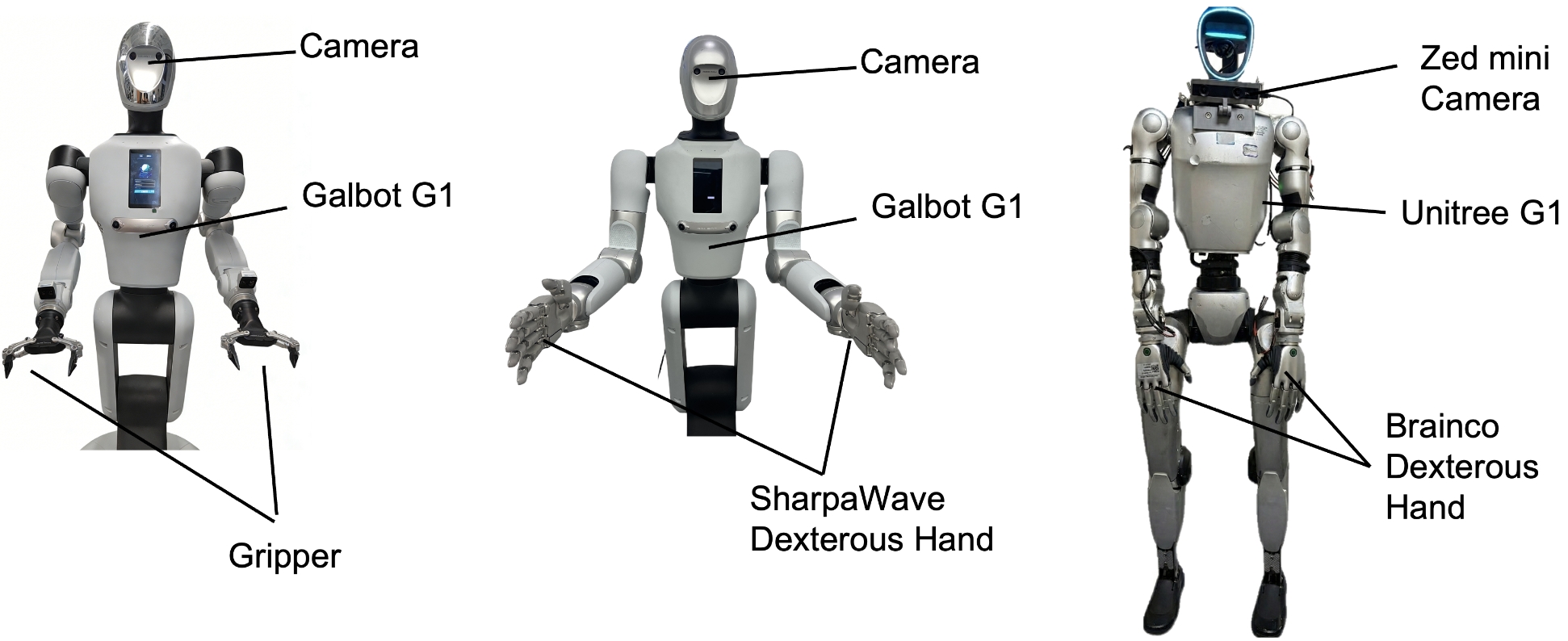}
    \caption{ Real-world robot platforms used in our physical experiments. From left to right:
    (1) Galbot G1 equipped with a standard two-finger parallel gripper for basic grasping tasks;
    (2) Galbot G1 fitted with the SharpaWave dexterous hand (22 DoF) for fine manipulation;
    (3) Unitree G1 mounted with the Brainco dexterous hand (10 DoF) and a Zed Mini camera.
    This multi-platform setup demonstrates the generalization capability of our LDA model across diverse robot morphologies and end-effectors.}
    \vspace{-2mm}
	\label{fig:real_setup}
\end{figure*}
\begin{figure*}[t]
    \centering
    \includegraphics[width=0.90\linewidth]{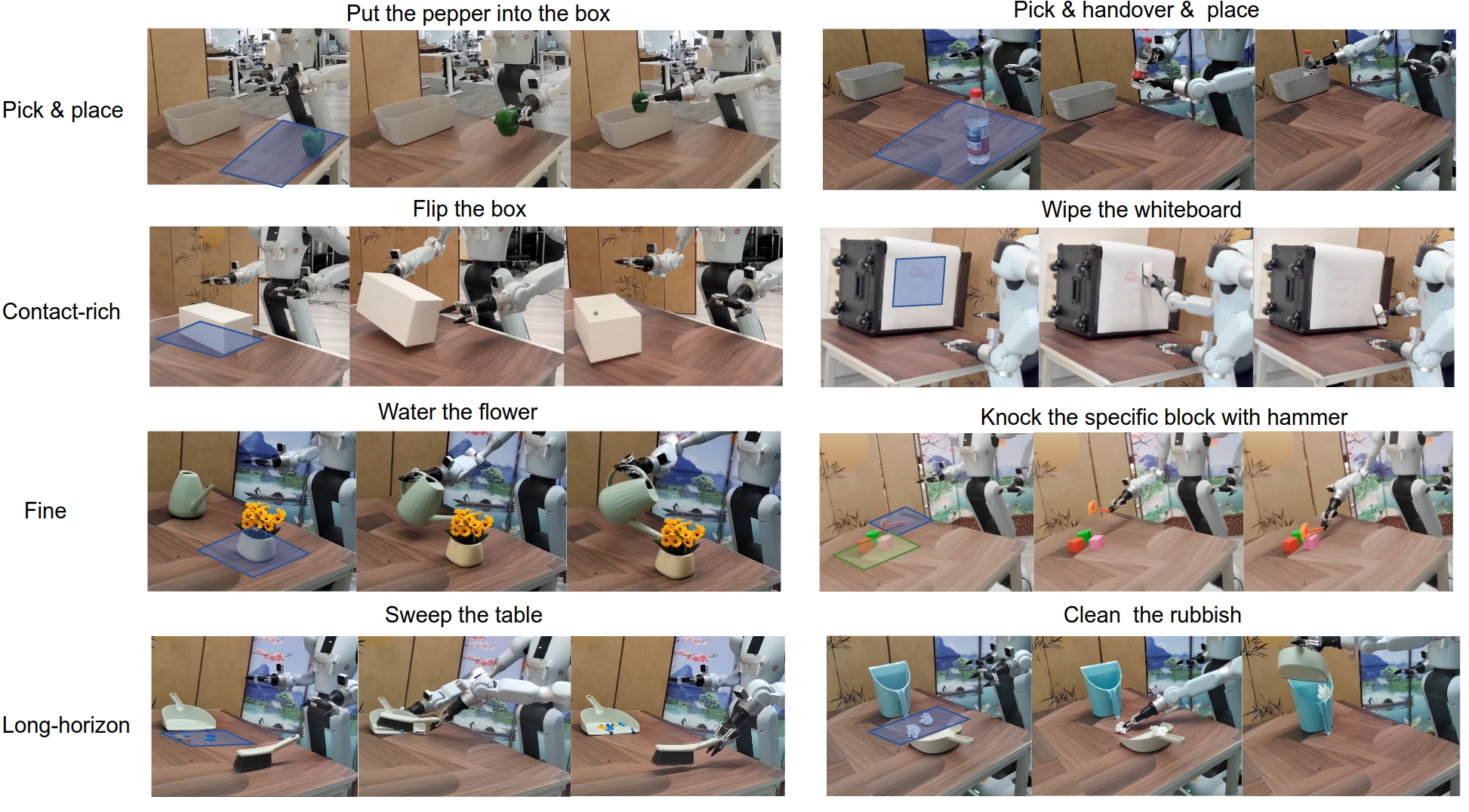}
    \caption{Task descriptions for the Galbot G1 robot equipped with a standard two-finger parallel-jaw gripper, spanning four manipulation categories.}
    \vspace{-2mm}
	\label{fig:task_overview}
\end{figure*}
\section{Details regarding real-world experiment}

\subsection{Real-world Setup.}
We conduct real-world experiments on two humanoid platforms: the Galbot G1 and the Unitree G1, as shown in Fig.~\ref{fig:real_setup}. The Galbot G1, with two 7-DoF arms, is equipped with two interchangeable end-effectors: two-finger parallel-jaw grippers and 22-DoF SharpaWave dexterous hands. The Unitree G1 uses 10-DoF BrainCo hands. In all real-robot configurations, the policy receives visual input only from an egocentric head-mounted camera, providing a first-person view of the workspace.

\subsection{Task description and evaluation protocol.}
To validate the effectiveness of our method on physical systems, we evaluate eight representative manipulation tasks involving single-arm, dual-arm coordination, tool use, and contact-rich interactions. 
For object generalization, movable objects are randomized within predefined spatial regions while several supporting objects (e.g., baskets, dustpans, and trash bins) remain fixed to isolate task-specific manipulation challenges rather than compounding errors from initial grasp failures. All experiments are conducted in-domain, and each trial is terminated after 200 seconds if unsuccessful.
Task success is defined using task-specific criteria such as successful object placement, execution of full procedural steps, or normalized scoring metrics for partial completion in long-horizon tasks.
We evaluate each task over independent trials. The corresponding training data volume and success criteria for each task are summarized in Table~\ref{tab:real_world_tasks}.

\begin{figure}[t]
    \centering    \includegraphics[width=\linewidth, keepaspectratio]{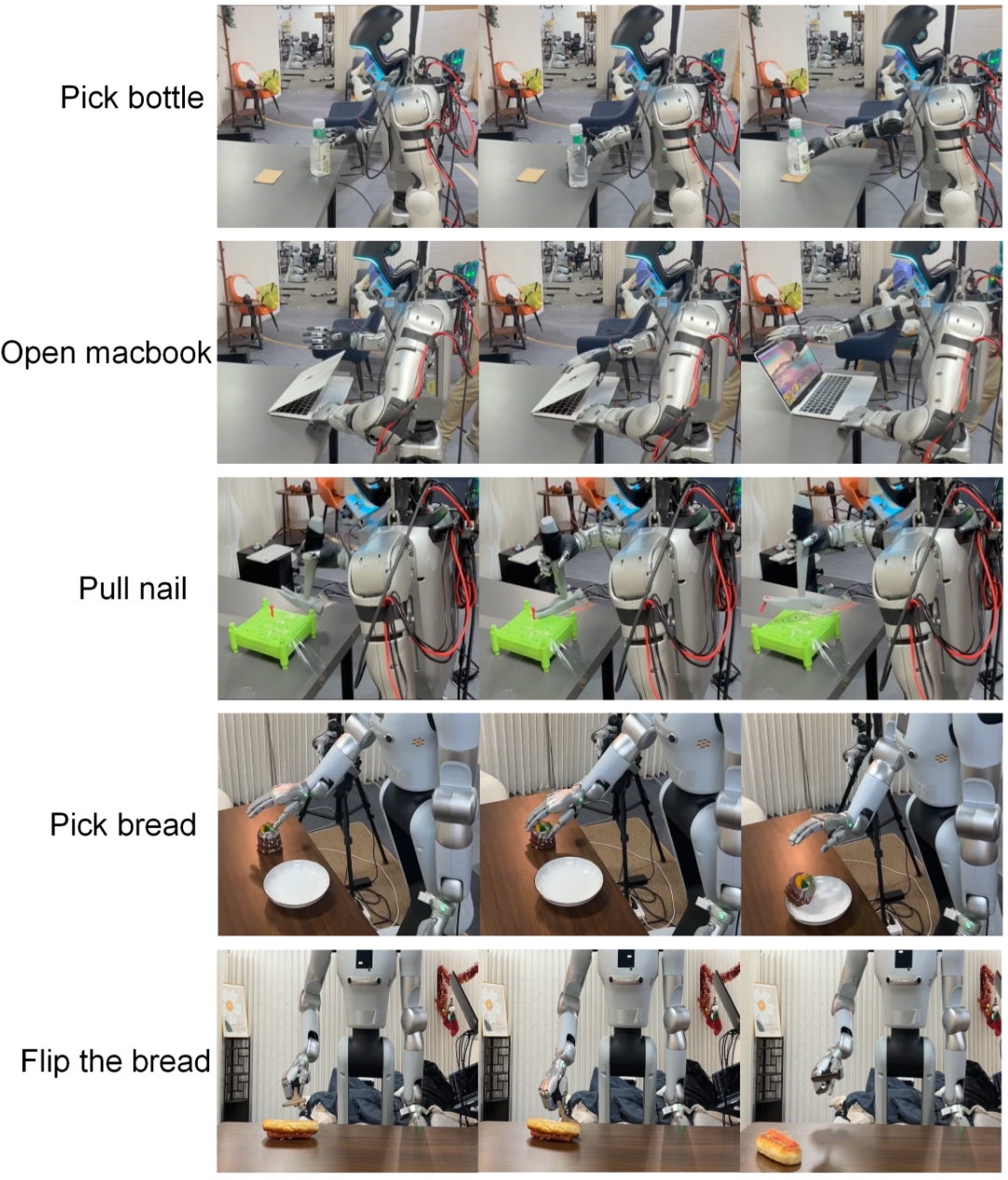}
    
    \caption{ Dexterous manipulation task descriptions across two robotic platforms. Top three rows: Unitree robot equipped with BrainCo hands performing bottle placement, MacBook opening, and nail extraction. Bottom two rows: Galbot robot utilizing SharpaWave hands executing bread placement and flipping tasks.}
    \vspace{-2mm}
	\label{fig:dex_demo}
\end{figure}

\begin{table*}[htbp]
\centering
\small
\setlength{\tabcolsep}{3.5pt}
\renewcommand{\arraystretch}{1.15}
\begin{tabular}{p{3.3cm} p{6.6cm} p{4.8cm}}
\toprule
\textbf{Task Abbreviation} & \textbf{Description} & \textbf{Test Protocol} \\
\midrule

Pick Vegetable 
& Pick a plastic pepper and place it into a basket using the left gripper. Pepper is randomized within a 15$\times$30 cm region. 
& 10 trials; success if placed in basket \\ \midrule

Handover 
& Left gripper grasps a bottle and passes it to the right gripper, which places it into a basket. Bottle randomized within 15$\times$30 cm. 
& 10 trials; success if placed in basket \\\midrule

Wipe Board 
& Use an eraser to remove marker writing from a whiteboard. Writing area randomized within 25$\times$40 cm. 
& 10 trials; scored from 0--5 based on cleaning completeness \\\midrule

Flip Box 
& Flip an upside-down storage box to upright using bimanual manipulation. Box randomized within 2$\times$4 cm. 
& 10 trials; success if fully flipped \\\midrule

Water Flower (pouring)
& Grasp a watering bottle and pour water into a flower pot. Pot randomized within 15$\times$15 cm. 
& 10 trials; success if pouring posture is achieved with spout above pot \\\midrule

Knock the block with a hammer (pnp2)
& Grasp a hammer with a very thin handle and then knock the specific block. Hammer randomized within 15$\times$15 cm. 
& 60 trials; success only if both the grasp and knock succeed. \\\midrule

Sweep Table 
& Sweep ten nails into a dustpan using a broom and dustpan. Nail positions randomized within 10$\times$25 cm. 
& 10 trials; success rate is computed as the proportion of nails collected in the dustpan. \\\midrule

Throw Rubbish 
& Pick paper balls, place them into a dustpan, and dump them into a trash can. Paper balls are randomized within a 20$\times$25 cm area.
& 10 trials; success rate is computed as the proportion of paper balls successfully dumped into the trash can.\\
\bottomrule
\end{tabular}
\caption{Real-world gripper manipulation for Galbot task configurations. All tasks are evaluated in-domain with a timeout of 200 seconds per trial.}
\label{tab:real_world_tasks}
\end{table*}

\subsection{More Analysis.}
To validate the efficacy of our proposed approach, we conducted a comprehensive comparison against two baseline policies: GR00T-N1.6~\cite{bjorck2025gr00t} and $\pi_{0.5}$~\cite{intelligence2025pi_}. Our method (LDA) demonstrates superior performance across all four evaluated categories: Pick \& Place, Contact-rich Manipulation, Fine Manipulation, and Long-horizon Manipulation.

\textbf{Performance on Basic Grasping Tasks}. In standard Pick \& Place scenarios, LDA achieves a dominant success rate, reaching 90.0\% on the ``handover" task, significantly outperforming $\pi_{0.5}$~\cite{intelligence2025pi_} (70.0\%) and nearly doubling the success rate of GR00T-N1.6~\cite{bjorck2025gr00t} (50.0\%). This indicates that our policy has learned a more robust grasping primitive and achieves better few-shot adaptation on the unseen Galbot robot, benefiting from larger-scale cross-embodiment learning.

\textbf{Robustness in Contact-Rich and Fine Manipulation}. The advantages of LDA become increasingly pronounced in tasks requiring precise dynamic interaction. 
In Contact-rich Manipulation, such as ``flip the box," LDA achieves a 60.0\% success rate compared to just 20.0\% for GR00T-N1.6~\cite{bjorck2025gr00t}. This suggests that LDA effectively models the complex contact dynamics required to manipulate objects without slippage or instability, whereas the baselines likely struggle with the discontinuous nature of the contact forces.
Similarly, in Fine Manipulation tasks like ``pouring," which demand continuous closed-loop feedback, our method sustains an 80.0\% success rate, surpassing the best baseline ($\pi_{0.5}$~\cite{intelligence2025pi_}) by 20 percentage points.

\textbf{Capabilities in Long-Horizon Planning}. 
The most striking distinction emerges in the Long-horizon Manipulation category. While baseline methods achieve moderate success on the relatively simple “sweep the table” task, they completely fail on the more complex “throw rubbish” task, registering a 0.0\% success rate. In stark contrast, LDA achieves a 35.0\% success rate, demonstrating robustness in multi-stage, temporally extended scenarios.
This performance gap reveals a fundamental limitation of existing approaches: their inability to manage compounding errors over long action sequences due to a lack of explicit dynamics modeling. LDA's success stems from its capacity to reason about the physical consequences of actions across time, maintain temporal consistency in latent states, and recover from intermediate deviations, capabilities that are essential for real-world, multi-step manipulation. Crucially, this advantage is rooted in LDA's dynamics-aware architecture, which aligns predicted visual features with underlying physical transitions and mitigates covariate shift through structured temporal modeling.
Collectively, these results validate that explicitly modeling latent dynamics is not merely beneficial but \emph{necessary} for reliable, generalizable robotic manipulation in complex, real-world settings.

\begin{table*}[htbp]
\centering
\small
\setlength{\tabcolsep}{4pt}
\renewcommand{\arraystretch}{1.2}
\begin{tabular}{p{2.4cm} p{6.6cm}  p{4.8cm}}
\toprule
\textbf{Task Abbreviation} & \textbf{Description}  & \textbf{Test Protocol} \\
\midrule

Pick Bottle
& Pick up a plastic bottle and place it onto a fixed target region using the right hand. Bottle position is randomized.

& 20 trials; success if bottle is upright and its base overlaps at least half of the target region \\
\midrule
Open MacBook
& Left hand stabilizes the base while the right hand opens the hinge by pushing the upper edge. Initial opening angle is randomized.

& 20 trials; success if opening angle exceeds 75\% of maximum \\
\midrule
Pull Nail
& Use a claw hammer held by the right hand to extract a nail from the surface. Hammer pose is randomized.

& 10 trials; scored with partial credit: 0.25 for locating, 0.5 for single-claw removal, 1.0 for full claw removal \\
\midrule
Pick Bread
& Pick a bread item and place it into a plate using the right hand. Three bread types are used with equal distribution.

& 10 trials; success if bread is placed into the plate \\
\midrule
Flip Bread
& Flip a long bread item using a spatula held by the right hand. Bread pose is randomized over a large region.

& 10 trials; 1.0 if flipped on first attempt, 0.5 if second, 0 otherwise \\

\bottomrule
\end{tabular}
\caption{Dexterous hand manipulation tasks and evaluation protocols.}
\label{tab:dexterous_hand_tasks}
\end{table*}

\textbf{Capabilities in Dexterous Manipulation.}
LDA consistently outperforms baselines on both low-DoF and high-DoF hands, with the performance gap becoming more pronounced as task difficulty and dexterity requirements increase.
For low-DoF hands, LDA already demonstrates strong robustness on tasks involving tool use and force-sensitive interactions. On \emph{Pick Bottle}, LDA achieves a 90\% success rate, substantially higher than $\pi_{0.5}$ (20\%) and GR00T-N1.6 (75\%). On \emph{Pull Nail}, which requires precise force direction and stable contact maintenance, LDA reaches 80\% success, while $\pi_{0.5}$ completely fails and GR00T-N1.6 achieves only 40\%. Notably, all methods perform well on \emph{Open MacBook}, suggesting that tasks with strong geometric affordances and limited contact ambiguity are less challenging even for baseline policies.
The advantage of LDA becomes even more evident with high-DoF hands, where action spaces are larger and control errors accumulate more easily. On \emph{Pick Bread}, LDA attains a 70\% success rate, outperforming GR00T-N1.6 (20\%) and $\pi_{0.5}$ (10\%). The gap further widens on \emph{Flip Bread}, a highly dexterous task requiring coordinated finger motion and continuous contact reasoning, where LDA achieves 90\% success while both baselines remain at only 10\%.
These results highlight LDA's superior ability for high-dimensional control and contact-rich dexterous manipulation. Unlike baseline methods that rely primarily on reactive policies, LDA benefits from dynamics-aware latent representations that capture fine-grained physical interactions over time. This enables more stable control, improved contact reasoning, and effective recovery from transient failures, capabilities that are critical for dexterous manipulation with complex, multi-DoF robotic hands.

\section{Details of EI-30k.}
\subsection{Data Processing Pipeline for Robot and Human Datasets}

To ensure consistency and usability across heterogeneous robot and human datasets, we design a standardized data processing pipeline that converts raw recordings into a unified representation suitable for effective learning of both policy and dynamics. The pipeline consists of three main stages: dataset standardization, coordinate alignment and cleaning, and post-processing for training.

\paragraph{Dataset Standardization}
All raw datasets are first converted into the common LeRobot~\cite{cadene2024lerobot} 2.1 format. This format includes:
\begin{itemize}
    \item \textbf{End-effector poses}: 6D position and orientation for both hands (human) or manipulators (robot);
    \item \textbf{Hand articulation}: 21-point MANO keypoints for human hands (when available) and binary or continuous gripper states for robots;
    \item \textbf{Camera parameters}: intrinsic and extrinsic matrices enabling reprojection across coordinate frames;
    \item \textbf{Task and temporal metadata}: task identifiers, episode boundaries and timestamps.
\end{itemize}

During this stage, all sequences are uniformly resampled to 10 Hz, and structured metadata files are generated to preserve the alignment between frames and their semantic annotations, ensuring temporal coherence and task-aware data organization for downstream  training.

After standardizing to LeRobot format, we implement an easy-to-use Dataset class for the following data processing pipeline to harmonize heterogeneous action data across diverse datasets. 

\begin{python}
class EmbodiedDataset:
    def __init__(self, dataset: str, eef_in_world: 
    int, has_mano: bool):
        self.dataset = dataset
        self.eef_in_world = eef_in_world  
        # 1 if wrist in world coordinates
        self.has_mano = has_mano
        self.eef_offset = {hand: np.eye(4) for hand
        in HAND_KEYS}
        self.eef_keys = HAND_KEYS

    def get_wrist(self, df: pd.DataFrame) -> 
    dict[str, np.ndarray]:
        pass

    def get_mano_or_gripper(self, df: pd.DataFrame):
        pass

\end{python}

\paragraph{Coordinate Alignment and Data Cleaning}

Human and robot datasets often employ inconsistent coordinate frame definitions. To unify them, particularly the end-effector (EEF) representations, we apply the following alignment and cleaning steps:

\begin{itemize}
    \item \textbf{End-effector coordinate alignment}: For each dataset, we define a canonical EEF frame (e.g., at the wrist or gripper center). All recorded hand or manipulator poses are transformed into this common frame using a dataset-specific rigid offset, estimated through geometric inspection or visual validation.
    
    \item \textbf{Camera motion decoupling}: For sequences captured in a moving camera frame, hand trajectories are reprojected into a fixed world coordinate system to eliminate artifacts caused by camera motion.
    
    \item \textbf{Keypoint standardization}: Human hand poses without native MANO keypoints are converted into the standard 21-point MANO representation, expressed relative to the aligned wrist frame.
    
    \item \textbf{Data validation}: Hand visibility is verified using an off-the-shelf detector; frames with occluded, truncated, or kinematically invalid hand data are discarded to ensure annotation reliability.
\end{itemize}

For robot datasets, we further normalize actuation signals: gripper widths are scaled to a consistent range (e.g., $[0, 1]$), and joint encodings are harmonized to match a unified kinematic convention.







\paragraph{Data Cleaning}
Textual annotations are unified into a structured format that explicitly describes the environmental context, per-hand actions (left/right), and high-level task objectives. When original annotations are inconsistent or missing, we leverage vision-language models to generate coherent, semantically aligned instructions.
Finally, all processed datasets are organized by agent type (human or robot) and accompanied by comprehensive metadata files detailing task definitions, episode boundaries, and dataset statistics. This standardized pipeline ensures a consistent, interoperable data representation across domains, enabling robust, scalable training of dexterous manipulation policies that generalize across embodiment and task complexity.





\begin{table}[htbp]
\centering
\label{tab:dataset_composition}
\resizebox{\columnwidth}{!}{
\begin{tabular}{llcc}
\toprule
\textbf{Data Type} & \textbf{Source / Sub-dataset} & \textbf{Duration (h)} \\ \midrule
\multirow{7}{*}{\textbf{Real-world Robot}} & Open X-Embodiment~\cite{o2024open} & 3000  \\
 & Agibot World~\cite{bu2025agibot_iros} & 3276  \\
 & RoboMIND~\cite{wu2025robomind} & 305  \\ 
 & Humanoid Everyday~\cite{zhao2025humanoideverydaycomprehensiverobotic} & ~30 \\
 & RoboCOIN~\cite{wu2025robocoin} & 500  \\
 & Galaxea~\cite{galaxea2025} & 500  \\
 & LET~\cite{LET2025} & 1000  \\ \midrule
 
\multirow{2}{*}{\textbf{Simulated Robot}} & InternData-A1~\cite{chen2025internvla} & 7433  \\
 & Behavior-1k~\cite{li2023behavior} & 1200  \\ \midrule

\multirow{12}{*}{\textbf{\shortstack[l]{Ego Human \\(w/ Action)}}} & Ego4D~\cite{grauman2022ego4d} & 3670  \\
 & Epic-Kitchens~\cite{damen2020epic} & 100  \\
 & Ego-Exo4D~\cite{grauman2024ego} & 1286  \\
 & SSV2~\cite{goyal2017something} &240  \\
 & EgoDex~\cite{hoque2025egodex} &830  \\
 & HOT3D~\cite{banerjee2025hot3d} &16  \\
 & HoloAssist~\cite{wang2023holoassist} &166  \\
 & OAKINK2~\cite{zhan2024oakink2} &6.5 \\
 & TACO~\cite{liu2024taco} &3.2  \\ 
 & HOI4D~\cite{liu2022hoi4d} &7.6  \\ 
 & ARCTIC~\cite{fan2023arctic} &2.3  \\ \midrule

\multirow{3}{*}{\textbf{\shortstack[l]{Ego Human \\(Actionless)}}} & Egocentric-10k~\cite{buildaiegocentric10k2025} & 10000  \\
 & RH20T-human~\cite{fang2023rh20tcomprehensiveroboticdataset} &100  \\
 & EgoMe~\cite{qiu2025egome} & 80  \\
  & Taste-Rob~\cite{zhao2025taste} &130 \\
 \midrule 
 & \textbf{Total} & \textbf{30k+}  \\ \bottomrule
\end{tabular}%
}
\caption{Composition of the Embodied Interaction Dataset (EI-30k). The dataset is categorized into four main types, aggregating over 30k hours of data.}
\end{table}

\begin{figure}[htbp]
    \centering
    \includegraphics[width=\linewidth]{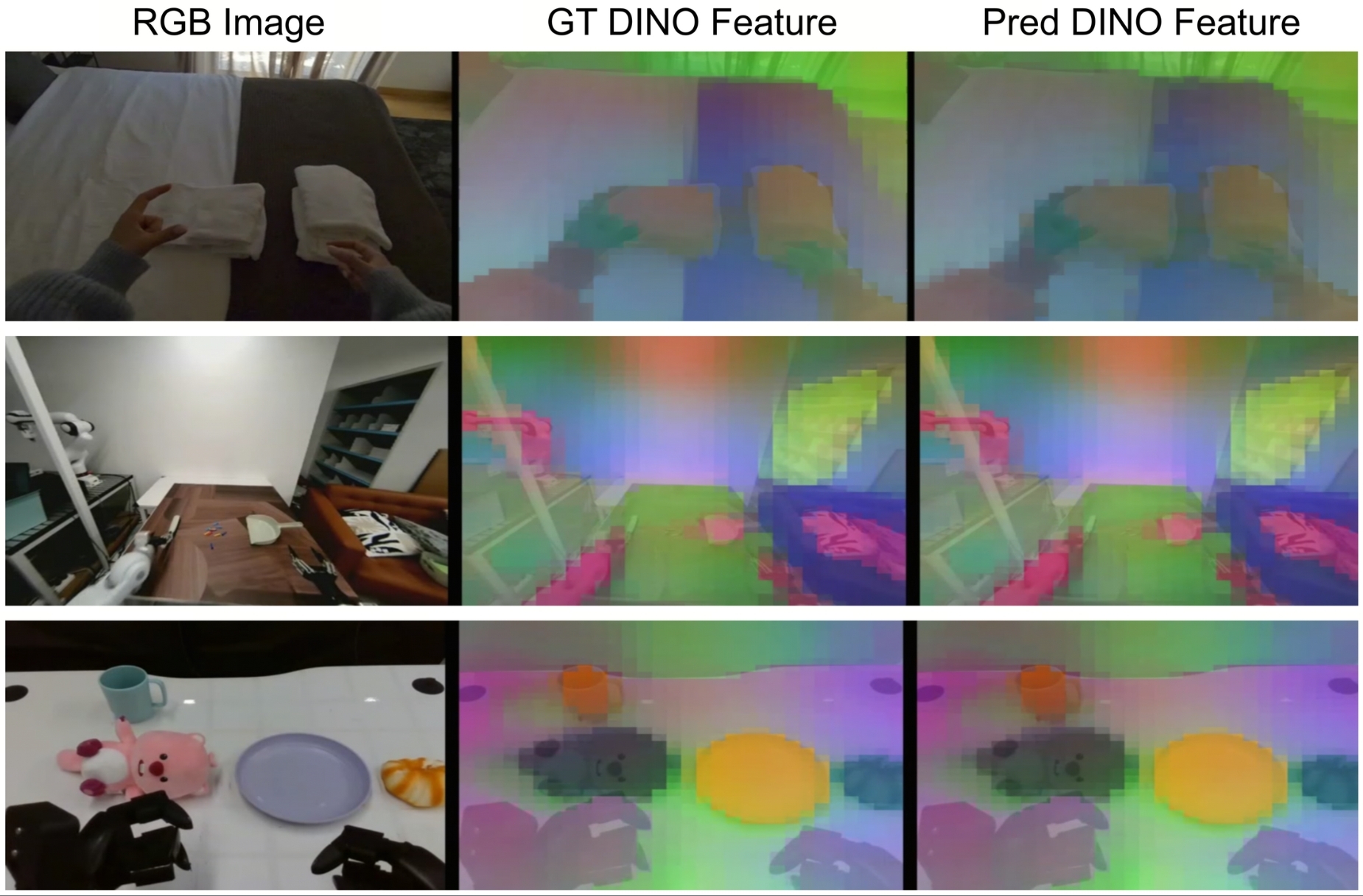}
    \caption{DINO Feature Prediction Visualization. Left column: Original RGB input images. Middle column: Ground-truth DINO features extracted by DINOv3~\cite{simeoni2025dinov3}. Right column: DINO features predicted by our model. }
    \vspace{-5mm}
	\label{fig:dino_and_rgb}
\end{figure}

\subsection{Data Composition.}
Our training data spans four complementary categories, totaling more than 30,000 hours of egocentric experience:

\setcounter{paragraph}{0}
\paragraph{Real-world Robot Data}  
This category includes large-scale physical robot execution logs. We primarily leverage \textit{Open X-Embodiment}~\cite{o2024open} and \textit{Agibot World}~\cite{bu2025agibot_iros} for general-purpose manipulation. To enhance hardware-specific capabilities, we incorporate \textit{Humanoid Everyday}~\cite{zhao2025humanoideverydaycomprehensiverobotic} for bipedal locomotion dynamics and \textit{Galaxea}~\cite{galaxea2025} for high-fidelity dexterous tasks. Additionally, we include \textit{RoboCOIN}~\cite{wu2025robocoin} despite its noisier action labels, as it provides valuable diverse environment explorations.

\paragraph{Simulated Robot Data}
To provide dense, noise-free supervision, we use high-quality simulated trajectories. The majority comes from \textit{InternData-A1}~\cite{chen2025internvla}, which offers large-scale automated generation of locomotion and basic manipulation sequences. \textit{Behavior-1k}~\cite{li2023behavior} further contributes long-horizon task demonstrations in simulated household environments, enabling the model to learn complex task hierarchies.

\paragraph{Egocentric Human Data with Actions}
This subset bridges human intent and robot-executable actions. We draw from large-scale datasets such as \textit{Ego4D}~\cite{grauman2022ego4d}, \textit{Epic-Kitchens}~\cite{damen2020epic}, \textit{Ego-Exo4D}~\cite{grauman2024ego} and \textit{SSV2}~\cite{goyal2017something} focusing on object-interaction segments. High-precision sources like \textit{EgoDex}~\cite{hoque2025egodex} and \textit{HOT3D}~\cite{banerjee2025hot3d} provide fine-grained 3D hand poses and contact information, critical for learning extrinsic dexterity.  

\paragraph{Egocentric Human Data without Actions}
Representing the largest source of visual diversity, this category consists of first-person observations. \textit{Egocentric-10k}~\cite{buildaiegocentric10k2025} serves as the primary source, covering a broad spectrum of daily activities. Additional datasets like \textit{RH20T-human}~\cite{fang2023rh20tcomprehensiveroboticdataset} and \textit{Taste-Rob}~\cite{zhao2025taste} contribute domain-specific visual priors. Although these trajectories lack explicit action labels, they provide a powerful self-supervised signal for learning world dynamics, visual affordances, and temporal structure.

\section{Details of Other Experiments}

\subsection{Action-Conditioned Attention Visualization.}
\label{app:action_attn}

We provide additional details on how action-conditioned attention maps are computed and interpreted.
Our visualization is based on the Diffusion Transformer (DiT)~\cite{Peebles2022DiT} backbone, where visual tokens and action embeddings interact through shared self-attention layers.

For a given observation, we extract attention maps from the middle transformer blocks, where high-level semantic and geometric information is most prominent.
Conditioned on an active action primitive (e.g., ``Push Right''), we compute the attention weights \( A_1 \), which quantify the influence of each spatial token on the predicted latent transition.
To establish a reference, we generate a baseline attention map \( A_2 \) by replacing the action embedding with a \emph{No-Op} (static) command.

We then compute the absolute difference:
\[
\Delta A = |A_1 - A_2|,
\]
which isolates attention changes induced purely by the action condition.
This subtraction effectively removes generic visual saliency (e.g., high-contrast edges or background objects) and highlights regions whose relevance emerges only when a specific action is applied.

As illustrated in Fig.~\ref{fig:attn_diff}, the resulting difference maps consistently emphasize contact regions, force application points, and anticipated motion trajectories.
For example, in the ``Push Right'' task, attention shifts toward the gripper-object contact interface and the direction of expected displacement.
This behavior demonstrates that the DiT dynamically re-weights visual tokens based on the physics implied by the action, rather than passively encoding static appearance.

\subsection{Visualization of Latent Forward Dynamics}
\label{app:dino_vis}

We provide additional qualitative visualizations to illustrate the forward dynamics learned by LDA.
These qualitative results complement the quantitative analysis and provide further evidence that LDA learns structured, dynamics-aware latent representations suitable for long-horizon reasoning and control.

\end{appendices}

\end{document}